%% file: main.tex
\PassOptionsToPackage{table,dvipsnames}{xcolor}
\documentclass[10pt]{article} %
\usepackage[accepted]{tmlr}

\usepackage{microtype}
\usepackage{xcolor,colortbl}
\definecolor{citecolor}{rgb}{0,0,0.5}
\usepackage[pagebackref,colorlinks=true,citecolor=citecolor]{hyperref}

\usepackage{booktabs}
\usepackage{multirow}
\usepackage{makecell}
\usepackage{float,subcaption}
\usepackage{wrapfig}
\usepackage{enumitem}

\input{def}
\title{Offline Learning and Forgetting for Reasoning with Large Language Models}

\author{\name Tianwei Ni$^1$\thanks{Work primarily done while Tianwei Ni was an intern at Amazon. Correspondence: Tianwei Ni (\texttt{twni2016@gmail.com}), Allen Nie (\texttt{anie@cs.stanford.edu}), Rasool Fakoor (\texttt{rasool.fakoor@gmail.com}).}, Allen Nie$^2$, Sapana Chaudhary$^2$, Yao Liu$^2$, Huzefa Rangwala$^2$, Rasool Fakoor$^2$ \\
        \addr $^1$Mila - Quebec AI Institute \& Université de Montréal, $^2$Amazon Web Services
}

\def\month{10}  %
\def\year{2025} %
\def\openreview{\url{https://openreview.net/forum?id=RF6raEUATc}} %

\begin{document}

\maketitle

\input{content}

{
\small
\bibliography{citation}
\bibliographystyle{tmlr}
}

\appendix
\input{appendix}

\end{document}

%% file: def.tex
\usepackage{algorithm}
\usepackage{algpseudocode}

\usepackage{amsmath}
\allowdisplaybreaks %
\usepackage{amssymb}
\usepackage{mathtools}
\usepackage{amsthm}
\usepackage{nicefrac}
\usepackage{thmtools}
\usepackage{bbm}
\usepackage{cancel}
\usepackage{centernot}
\usepackage{siunitx}
\usepackage{tikz}
\usetikzlibrary{decorations,arrows,shapes}

\usepackage{listings}
\usepackage{xspace}
\makeatletter
\DeclareRobustCommand\onedot{\futurelet\@let@token\@onedot}
\def\@onedot{\ifx\@let@token.\else.\null\fi\xspace}

\def\eg{\emph{e.g}\onedot}

\makeatother

\newcommand{\E}[2]{\mathbb{E}_{#1}{\left[#2\right]}}

\newcommand{\D}{\mathcal{D}}

\newcommand{\x}{\mathbf{x}}
\newcommand{\y}{\mathbf{y}}

\newcommand{\defeq}{\mathrel{\mathop:}=}

\usepackage{pifont}%

%% file: content.tex
\begin{abstract}

Leveraging inference-time search in large language models has proven effective in further enhancing a trained model's capability to solve complex mathematical and reasoning problems. However, this approach significantly increases computational costs and inference time, as the model must generate and evaluate multiple candidate solutions to identify a viable reasoning path. To address this, we propose an effective approach that integrates search capabilities directly into the model by fine-tuning it on \textit{unpaired} successful (\textit{learning}) and failed reasoning paths (\textit{forgetting}) derived from diverse search methods. A key challenge we identify is that naive fine-tuning can degrade the model's search capability; we show this can be mitigated with a smaller learning rate. Extensive experiments on the challenging Game-of-24 and Countdown arithmetic puzzles show that, replacing CoT-generated data with search-generated data for offline fine-tuning improves success rates by around 23\% over inference-time search baselines, while reducing inference time by 180$\times$. On top of this, our learning and forgetting objective consistently outperforms both supervised fine-tuning and preference-based methods.\footnote{Code is open-source at \url{https://github.com/twni2016/llm-reasoning-uft}.}\looseness=-1

\end{abstract}

\section{Introduction}

\begin{wrapfigure}{r}{0.42\textwidth}
\vspace{-4.5em}
  \begin{center}
\includegraphics[width=0.92\linewidth]{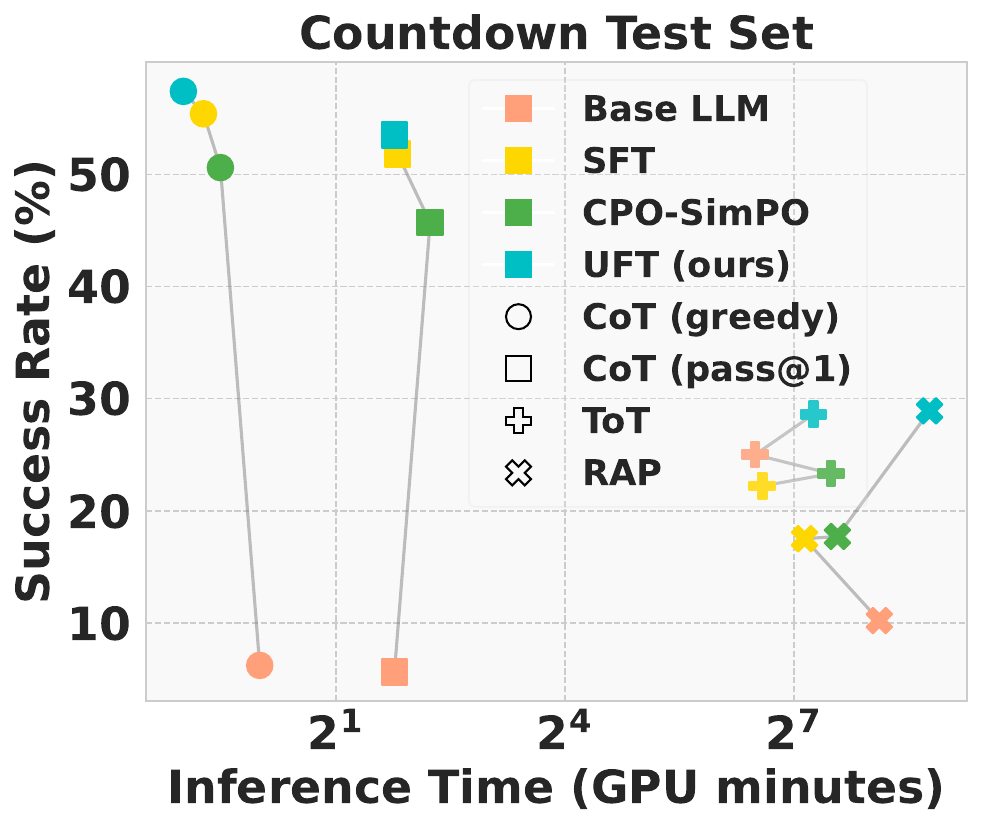}
  \end{center}
  \vspace{-1.5em}
  \caption{\small Trade-off between inference time and success rate in challenging arithmetic puzzles, \textit{Countdown}. 
  We find that fine-tuning on \textbf{CoT-style data from diverse reasoners} substantially enhances CoT inference over the base LLM (Qwen2.5-Math 7B) while preserving \textit{inference-time search} (ToT~\citep{yao2023tree}, RAP~\citep{hao2023reasoning}) using \textbf{a smaller learning rate}.
  Among fine-tuning methods including SFT and preference optimization, \textbf{our method (UFT)} achieves the best CoT and search performance.
  \looseness=-1}
\label{fig:teaser}
  \vspace{-2.5em} %
\end{wrapfigure}

The last few years has seen the rapid development of large language models (LLMs)~\citep{vaswani2017attention,achiam2023gpt} and their applications to a diverse set of tasks.
In particular, these LLMs have been tested on challenging benchmarks requiring high-level reasoning across domains such as mathematics~\citep{glazer2024frontiermath}, abstract reasoning~\citep{chollet2024arc}, code generation~\citep{zhuo2024bigcodebench}, and science~\citep{mitchener2025bixbench}.\looseness=-1

Chain-of-Thought (CoT) and \textit{inference-time search} (search) have been proposed to enhance LLM generalization to reasoning problems at test time. 
Specifically, given a task description and input $\x$, CoT~\citep{wei2022chain} outputs a single reasoning path $\hat \y$ via greedy decoding, or samples multiple paths to compute average performance. 
In contrast, search-based methods~\citep{lightman2023let,yao2023tree,snell2024scaling} explicitly generate intermediate reasoning (possibly with multiple passes) before extracting $\hat \y$ via a reward model or self-evaluation. 
Despite inference-time search's effectiveness, it comes with greater computational costs (e.g., \autoref{fig:teaser} and \citet{chen2024tree, chen2024not}).\looseness=-1

A beneficial outcome of utilizing inference-time sampling strategies, such as CoT and search-based methods, is the generation of rich datasets. These datasets can subsequently be fed back into the model for further fine-tuning, potentially enhancing the base model's overall performance.
Prior work has primarily focused on distilling the knowledge from \textit{successful} reasoning paths, drawing from CoT-generated data~\citep{zelikman2022star,yuan2023scaling,singh2023beyond}, paths from search trees~\citep{feng2023alphazero,tian2024toward,zhang2024rest}, or full search traces~\citep{gandhi2024stream,lehnert2024beyond}. 
However, in many complex problems, failed (reasoning) paths far outnumber successful ones, representing an underutilized resource. While potentially valuable, these failed paths are often unpaired, making them difficult to use in preference optimization~\citep{rafailov2023direct}. 
Diverse search algorithms further enrich data coverage and reasoning quality but produce heterogeneous structures (e.g., trees) unlike linear CoT traces.
This raises a natural question: \textit{can we make full use of both successful and failed reasoning paths, from both CoT and search-based methods, to improve LLM reasoning?}\looseness=-1

In this paper, we propose a straightforward method consisting of three phases designed to enhance the reasoning capabilities of large language models: data generation, fine-tuning, and evaluation. First, we collect heterogeneous reasoning data from both CoT and the search-based methods, and convert the data into a unified CoT-style path format. Next, we fine-tune the base LLM by \textit{learning} successful paths with the supervised fine-tuning (SFT) and \textit{forgetting} unpaired failed paths with the unlikelihood loss.
We refer to this objective as \textit{unlikelihood fine-tuning (UFT)}.
Finally, we evaluate the fine-tuned model on test sets using both CoT and inference-time search, with a primary focus on efficient CoT inference.
We demonstrate the effectiveness of our three-phase pipeline on two arithmetic puzzle domains: Game-of-24 and Countdown, which are widely used as testbeds for explicit search~\citep{yao2023tree,gandhi2024stream}. We summarize our key findings below, using Qwen2.5-Math 1.5B and 7B~\citep{yang2024qwen2} as base LLMs:\looseness=-1
\begin{enumerate}[leftmargin=12pt,itemsep=0.8pt,parsep=0pt,topsep=0pt,partopsep=0pt]
    \item \textbf{Data quality drives performance:} On Countdown, replacing CoT data with high-quality classic search data boosts CoT performance from $33.5\%$ to $57.1\%$, significantly outperforming the inference-time search baseline ($25\%$) with a $180\times$ reduction in inference cost.\looseness=-1
    \item \textbf{Unpaired forgetting reliably improves performance:} Incorporating failed paths using UFT improves CoT performance by $1$–$2\%$ on average, with up to $7\%$ gains in the best case. In contrast, preference-based methods like CPO-SimPO~\citep{xu2024cpo-simpo}  often underperform SFT due to the paired format.\looseness=-1
    \item \textbf{Controlled learning rate helps maintain search capability:} Naive fine-tuning can quickly degrade a model’s search ability. A substantially reduced learning rate helps prevent this, making it a simple but crucial step when fine-tuning with CoT-style data.\looseness=-1
\end{enumerate}

\section{Related Work}
\vspace{-0.5em}

\paragraph{LLM inference-time search.} 
Inference-time search has shown remarkable success across reasoning~\citep{wang2022self,stechly2024chain,hao2024llm,snell2024scaling} and planning tasks~\citep{valmeekam2023planning,zheng2024natural,nova2024palnningBenchamrk}.
Approaches include best-of-n sampling~\citep{nakano2021webgpt,wang2022self}, multi-agent debates~\citep{du2023improving}, and iterative correction via predefined rules~\citep{bai2022constitutional} or self-refinement~\citep{shinn2023reflexion,madaan2024self}. 
A key direction, which this paper focuses on, integrates \textit{classic search algorithms} with LLMs, employing BFS~\citep{yao2023tree,xie2023self}, DFS~\citep{qin2023toolllm}, A*~\citep{zhuang2023toolchain,meng2024llm,koh2024tree} and MCTS~\citep{hao2023reasoning,zhou2023language,zhao2024large,xie2024monte,gao2024interpretable}. 
While effective, inference-time search methods remain computationally expensive: self-refinement requires multiple expensive passes, multi-agent approaches demand substantial GPU memory, and search trees grow exponentially with depth~\citep{chen2024tree}.\looseness=-1

\vspace{-0.5em}
\paragraph{LLM policy distillation from synthetic data.} 
Synthetic data, generated by classic algorithms or LLMs, is widely used for SFT in reasoning tasks to mitigate the scarcity of human-annotated datasets~\citep{liu2024best}. However, despite its abundance, synthetic responses may fail to solve the task. To address this, most prior work distills reasoning knowledge to policy~\citep{hinton2015distilling} by selecting only correct responses. 
Policy distillation strategies can be categorized based on whether the source and target data originate from CoT or search algorithms. 
CoT-to-CoT distillation learns CoT reasoning from correct CoT paths filtered by rewards, widely used in practice~\citep{zelikman2022star,zelikman2024quiet,uesato2022solving,yuan2023scaling,gulcehre2023reinforced,singh2023beyond}. 
Search-to-search distillation either refines the proposal policy in tree search by learning from expert behavior~\citep{feng2023alphazero,tian2024toward,zhang2024rest} and preference pairs~\citep{chen2024advancing}, or learns a meta-policy by imitating entire search traces~\citep{yang2022chain,gandhi2024stream,lehnert2024beyond}.
Notably, \citet{zhang2024chain} performs search-to-CoT distillation using preference pairs from Tree-of-Thought without a verifier. 
In comparison, we focus on $\{$CoT, search$\}$-to-CoT distillation from diverse reasoners and distill \textit{unpaired} positive and negative data to the policy, labeled by a rule-based verifier.\looseness=-1

\vspace{-0.5em}
\paragraph{Offline fine-tuning with negative data.} 
Negative data has been extensively studied in \textit{LLM safety}, where the goal is to unlearn harmful content using objectives such as Gradient Ascent and unlikelihood training~\citep{welleck2019neural,keskar2019ctrl,yao2023large,zhang2024negative}. 
In preference optimization, negative examples are used to increase the \textit{relative} likelihood of preferred responses in \textit{paired} data $(\x,\y^+,\y^-)$, commonly used in alignment~\citep{rafailov2023direct,yuan2023rrhf,zhao2023slic,hong2024orpo,xu2024contrastive,meng2024simpo} and reasoning~\citep{pal2024smaug,pang2024iterative,zhang2024chain,setlur2024rl,chen2024advancing}. However, this paired format discards unpaired positives or negatives, reducing data efficiency. KTO~\citep{ethayarajh2024kto} addresses the limitation of relative likelihood~\citep{tuan2024gradient} by learning from unpaired data, although it requires a reference model.
In LLM reasoning, negative data is also used to train a reward model~\citep{cobbe2021training,uesato2022solving,lightman2023let,feng2023alphazero,zhang2024rest,hosseini2024v} for response re-ranking and search during inference. However, this does not inherently improve the base model as a policy. 
Instead of requiring preference pairs or reference models, we adopt unlikelihood training~\citep{welleck2019neural} to directly forget failed paths as an auxiliary loss.
Finally, concurrent work~\citep{wang2025reasoning} highlights the challenges of unlearning in reasoning tasks, namely forgetting both incorrect reasoning traces and final answers while retaining overall reasoning ability.
\looseness=-1 

\section{Preliminaries}
\vspace{-0.5em}

\paragraph{Reasoning task as an MDP.} We formalize a reasoning task as a token-level Markov decision process (MDP)~\citep{sutton1998reinforcement} $(\mathcal X, \mathcal Y, R, T)$. The initial state $s_0 = \mathbf{x} \in \mathcal X$ consists of a tokenized sequence representing the task description and input. An action $y_t \in \mathcal Y$ is chosen based on the current state $s_t$, which is the sequence of all previous tokens: $s_t=(s_0,y_{0:t-1})$. A terminal state $s_T$ is reached upon generating a special end-of-sequence token. The ground-truth reward function $R$ is a rule-based verifier that checks the correctness of $s_T$, yielding $r=R(s_T) \in \{0,1\}$, where $0$ means failure and $1$ means success. The goal is to find a policy $\pi$ that generates an \textit{action sequence} $y_{0:T-1}$ (denoted as $\mathbf{y}$) which maximizes the reward: 
$
\max_{y_{0:T-1}} R(s_T) = \max_{\y} R(\x, \y)
$. We refer to the terminal state $(\x, \y)$ as a (reasoning) \textit{path}.

\vspace{-0.5em}
\paragraph{LLM reasoners.} We use an LLM policy $\pi_\theta$, parameterized by $\theta$, along with a search algorithm $f$ as the reasoner. We denote this LLM reasoner as $f(\pi_\theta)$ and focus on three popular LLM reasoners~\citep{hao2024llm}:
(1) CoT~\citep{wei2022chain} that uses greedy search, (2) Tree-of-Thought (ToT)~\citep{yao2023tree} that uses beam search, (3) Reasoning-via-Planning (RAP)~\citep{hao2023reasoning} that uses MCTS~\citep{kocsis2006bandit}.\looseness=-1

CoT directly outputs one (via greedy decoding) or multiple reasoning paths (under non-zero temperature) without intermediate search process\footnote{CoT's performance is often measured by the \textit{average} success rate over generated reasoning paths, also known as the \textbf{pass@1} metric~\citep{chen2021evaluating,guo2025deepseek}, as used in this work.}. 
On the contrary, ToT and RAP are \textit{inference-time search} methods, first constructing a search tree of reasoning paths before selecting a final path $(\x, \y)$ as the answer. Their performance is evaluated based on the success rate of the selected reasoning path.  Please see \autoref{sec:llm_reasoners} for more details.

\section{Fine-tuning on Unpaired Correct and Failed Paths from Diverse Reasoners}

\begin{figure}[t]
\vspace{-3em}
    \centering
    \includegraphics[width=0.65\linewidth]{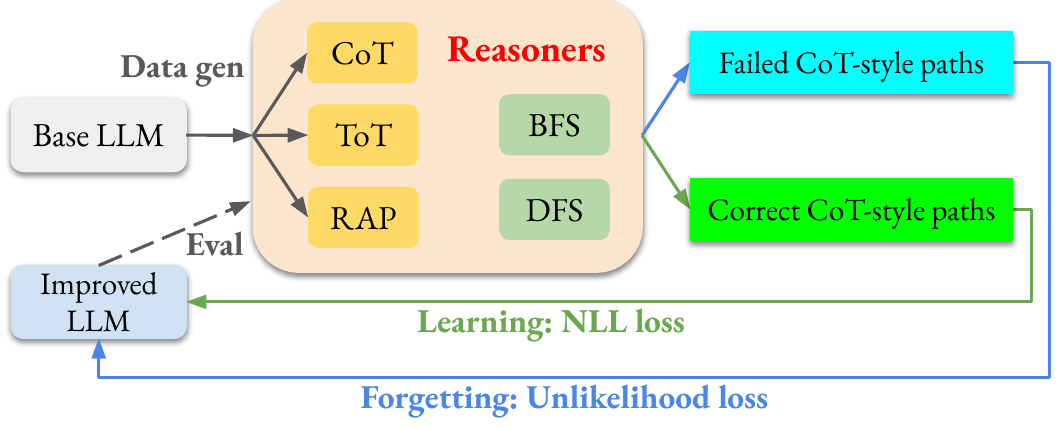}
    \vspace{-1em}
    \caption{\textbf{Our method for reasoning tasks.} We first generate synthetic reasoning data using multiple LLM reasoners (e.g., CoT, ToT, RAP) and classic algorithms (e.g., BFS, DFS). This data is converted to a unified CoT-style format and labeled as correct or failed by a ground-truth verifier to form an unpaired dataset. We then fine-tune the base LLM with negative log-likelihood (NLL) loss on correct paths and unlikelihood loss on failed paths, which we refer to as \textit{unlikelihood fine-tuning} (UFT). Finally, we evaluate the improved LLM with multiple LLM reasoners on a test set.\looseness=-1}
    \label{fig:pipeline}
    \vspace{-1.5em}
\end{figure}

In this section, we describe our approach to solving a reasoning task, as outlined in \autoref{fig:pipeline}. Our approach involves three stages: (1) generating reasoning data from diverse reasoners, (2) offline fine-tuning the base LLM from \textit{unpaired} correct and failed paths, and (3) evaluating it with reasoning algorithms.

\subsection{CoT-style Data Generation Using LLM and Classic Reasoners}

For a given reasoning task, we generate synthetic reasoning data from multiple reasoners. LLM reasoners $f(\pi_\theta)$ share a common base policy $\pi_\theta$ but differ in their search algorithms $f$, which include CoT and inference-time search methods such as ToT and RAP. Optionally, reasoning paths can also be generated using classical search algorithms (\textit{classic reasoners}) such as DFS and BFS, which rely on an external verifier rather than LLMs.\looseness=-1 

For search-based methods, we extract all root-to-leaf paths from the search tree -- rather than only the final selected path -- to follow the format of \textit{CoT-style paths}.
This enables (1) unifying heterogeneous search traces into a common format as training data, and (2) efficient inference, as CoT inference is fastest.
We aggregate CoT-style paths from all reasoners to compose the training dataset $\D$:
\begin{equation}
\D = \{(\x, \y, r) \mid f\in \mathcal F, \x \in \mathcal X, \y \in f(\pi_\theta)(\x), r = R(\x, \y)\},
\end{equation}
where  $\mathcal F$ is the set of considered reasoners, and $f(\pi_\theta)(\x)$ is the set of all reasoning paths produced by reasoner $f(\pi_\theta)$. Note that classic reasoners $f$ do not depend on $\pi_\theta$. 
Each path is labeled by a verifier as correct $(r=1)$ or failed $(r=0)$, splitting $\D$ into two unpaired datasets: the correct dataset $\D^{+} =\{(\x, \y) \mid (\x,\y, 1) \in \D\}$ and the failed dataset $\D^-= \{(\x, \y) \mid (\x,\y, 0) \in \D\} $. 

\subsection{Unlikelihood Fine-Tuning on Unpaired Correct and Failed Paths}
\label{sec:uft}
While prior work typically fine-tunes LLMs on correct CoT-generated paths~\citep{yuan2023scaling,singh2023beyond} or on preference pairs of correct and failed responses~\citep{pang2024iterative,zhang2024chain}, we extend these approaches in two directions: (1) incorporating diverse reasoners beyond CoT to \textit{augment} the training data, (2) fine-tuning the model to avoid and forget failed reasoning patterns using \textit{only} failed examples. Unlike most preference-based methods, our approach leverages \textit{unpaired} positive or negative data, without requiring success-failure pairs for the same question.

\vspace{-0.5em}
\paragraph{Learning to follow correct reasoning paths.}
First, we consider the negative log-likelihood (NLL) loss, also known as supervised fine-tuning (SFT)~\citep{ouyang2022training,cen2025bridging}, on correct reasoning paths collected from multiple reasoners:

\vspace{-0.5em}
\begin{equation}
\label{eq:nll}
\min_\theta -\E{(\x,\y,r)\sim\D}{\mathbf{1}(r = 1) \log \pi_\theta(\y\mid \x)} = -\E{(\x,\y^+)\sim\D^+}{ \log \pi_\theta(\y^+\mid \x)} \defeq J_{\text{NLL}}(\theta; \D^+).
\end{equation}

Examining the gradient of the NLL objective (\autoref{eq:nll}), $ -\E{(\x,\y,r)\sim\D}{\mathbf{1}(r = 1) \nabla \log \pi_\theta(\y\mid \x)}$, reveals its connection to REINFORCE~\citep{williams1992simple} with a binary (indicator) reward function, as noted in prior work~\citep{zelikman2022star,gulcehre2023reinforced,singh2023beyond}\footnote{This resemblance holds when $\D$ is close to on-policy data~\citep{fakoor20a}, e.g., CoT-generated, but may break down as model weights shift during fine-tuning or if $\D$ includes search-derived data.\looseness=-1}.\looseness=-1

\vspace{-0.5em}
\paragraph{Learning to avoid and forget incorrect reasoning paths.} While NLL encourages the model to learn correct reasoning paths, it can only leverage the limited number of available successful trajectories and discards a large number of failed ones -- which often vastly outnumber the correct ones (e.g.,  $|\D^-| / |\D^+|$ ranges from 5 to 400 in our experiments). To address the limitation of NLL, we propose leveraging failed trajectories by incorporating the \textbf{unlikelihood} (UL) loss~\citep{welleck2019neural} as an additional objective. The UL loss enables the model to learn to avoid (i.e., forget) failed reasoning paths, effectively utilizing the large set of failed trajectories.

\vspace{-0.5em}
\begin{equation}
\label{eq:ul}
\min_\theta J_{\text{UL}}(\theta; \D^-) \defeq -\E{(\x,\y^-)\sim \D^-}{\log (1- \pi_\theta(\y^- \mid \x))}.
\end{equation}
This objective explicitly reduces the probability of incorrect paths under the model.
Unlike Gradient Ascent (GA) unlearning~\citep{yao2023large}\footnote{The GA objective $\min_\theta \E{(\x,\y^-)\sim \D^-}{\log\pi_\theta(\y^-\mid \x)}$ is shown to be unstable to optimize~\citep{zhang2024negative,zhang2024safe}, as it has an unbounded optimum ($-\infty$) and a divergent gradient near the optimum. We provide results for GA in \autoref{app:unlearning} for completeness. }, UL objective has a smooth optimization landscape, with gradient: 
\vspace{-0.5em}
\begin{equation}
\label{eq:ul_grad}
\nabla J_{\text{UL}}(\theta)  = \E{(\x,\y^-)\sim \D^-}{\frac{\pi_\theta(\y^- \mid \x)}{1- \pi_\theta(\y^- \mid \x)}\nabla \log \pi_\theta(\y^- \mid \x)}.
\end{equation}
The gradient implies convergence to the stationary point where $\pi_\theta(\y^- \mid \x) = 0, \forall (\x,\y^-)\in \D^-$. 
From a REINFORCE view, \autoref{eq:ul_grad} imposes an \textit{adaptive penalty} of $-\frac{\pi_\theta(\y^- \mid \x)}{1- \pi_\theta(\y^- \mid \x)}$, penalizing higher-probability failures more strongly and thus efficiently suppressing wrong paths.\looseness=-1

\vspace{-0.5em}
\paragraph{Combining learning and forgetting: unlikelihood fine-tuning (UFT).}
Our final objective UFT combines NLL (\autoref{eq:nll})
and UL losses (\autoref{eq:ul}) with a coefficient $\alpha \in (0,1)$: 
\begin{equation}
\label{eq:obj}
\min_\theta J(\theta; \D, \alpha) \defeq (1-\alpha) J_{\text{NLL}}(\theta;\D^+) + \alpha J_{\text{UL}}(\theta; \D^-).
\end{equation}
Since correct paths provide explicit solutions while wrong paths only rule out alternatives, we treat UL as an \textit{auxiliary} loss by setting $\alpha$ close to 0.  
Moreover, as correct and failed paths may be drawn from the same search trees, they often share prefixes, introducing conflicts when $\alpha$ is large, as also noted in preference-based methods~\citep{zhang2024chain}. For example, suppose a prefix sequence $(\x, y_{0:t})$ ($t< T-1$) appears in both $\D^+$ and $\D^-$ with equal frequency. The gradient of \autoref{eq:obj} w.r.t. this prefix is 
$\left(\frac{\alpha}{1- \pi_\theta(y_{0:t} \mid \x)}-1\right)\nabla \log \pi_\theta(y_{0:t} \mid \x)
$, which has a stationary point at  $\pi_\theta(y_{0:t} \mid \x) = 1- \alpha$. Thus, $1-\alpha$ sets the desired probability of the shared prefix, further motivating the need for a small $\alpha$ to avoid conflicting objectives.  
Unlike prior work~\citep{zhang2024chain,setlur2024rl,chen2024advancing} that identify high-credit steps with more computation, we sidestep this with an auxiliary loss without preference pairs.  

\subsection{Practical Algorithm}
We summarize our method in \autoref{algo}. For a given reasoning task, we partition the initial state set $\mathcal X$ into training, validation and test sets, namely, $\mathcal X_{\text{train}}, \mathcal X_{\text{valid}}, \mathcal X_{\text{test}}$. Each subset has distinct input cases with follows the same instruction template. 
Our approach assumes access to a ground-truth verifier in $\mathcal X_{\text{train}}$ (for generating training data) and $\mathcal X_{\text{valid}}$ (for checkpoint selection), but not in  $\mathcal X_{\text{test}}$ on which the final evaluation is conducted. 
\begin{algorithm}[H]
\small
   \caption{Fine-tuning on unpaired correct and failed paths from diverse reasoners}
   \label{algo}
\begin{algorithmic}[1]
\Require Reasoning task $(\mathcal X, \mathcal Y, R, T)$, base LLM $\pi_\theta$, set of reasoners $\mathcal F$, number of epochs $E$, batch size $B$, learning rate $\eta$, UL coefficient $\alpha$ (close to 0)
\State Generate CoT-style training data from various reasoners: $
\D = \{(\x, \y, r) \mid f\in \mathcal F, \x \in \mathcal X_{\text{train}}, \y \in f(\pi_\theta)(\x), r = R(\x, \y)\}$, and split $\D$ into correct data $\D^+$ and failed data $\D^-$. 
\State Fine-tune the LLM with UFT (\autoref{eq:obj}) for $E|\D^+| / B$ steps. In each step, sample batches $\mathcal{B}^+ \sim \D^+$ and $\mathcal{B}^- \sim \D^-$ of size $B$ each, and update:
\begin{equation}
\theta \gets \theta - \eta [(1-\alpha) \nabla J_{\text{SFT}}(\theta;\mathcal B^+) + \alpha \nabla J_{\text{UL}}(\theta; \mathcal B^-)].
\end{equation}
\State Evaluate the fine-tuned LLM $\pi_\theta$ on test cases by collecting $\D' = \{(\x, \y, r) \mid f\in \mathcal F, \x \in \mathcal X_{\text{test}}, \y \in f(\pi_\theta)(\x), r = R(\x, \y)\}$. 
\end{algorithmic}
\end{algorithm}
\vspace{-0.5em}

\section{Experiments}
\vspace{-0.5em}

We evaluate our approach on two arithmetic puzzle tasks. 

\paragraph{Game-of-24.} 
In Game-of-24\footnote{\url{https://en.wikipedia.org/wiki/24_(puzzle)}}, the goal is to use basic arithmetic operations (\texttt{+-*/}) and parentheses to combine four input numbers to obtain the number $24$. Each input number can only be used once, which requires exactly three high-level reasoning steps. 
We follow the step-by-step response format from ToT~\citep{yao2023tree}. 
For example, given numbers \texttt{2 9 10 12}, a correct reasoning path is \texttt{12 * 2 = 24 (left: 9 10 24),
10 - 9 = 1 (left: 1 24),
24 * 1 = 24 (left: 24),
Answer: (12 * 2) * (10 - 9) = 24}. Each high-level step must include the remaining numbers after the operation. We implement a process-based verifier~\citep{uesato2022solving} that checks both the intermediate steps and the final answer. For evaluation, we use the same test set from \citet{yao2023tree}, consisting of $100$ cases. For training and validation, we use $900$ distinct cases, considered easier than the test set based on the human performance.\looseness=-1

\vspace{-0.5em}
\paragraph{Countdown.} The Countdown game~\citep{gandhi2024stream} extends Game-of-24 by requiring the use of four integers to obtain a specified target number beyond 24. All other rules remain the same including the step-by-step format. 
Following \citet{gandhi2024stream}, we randomly generate 500k training cases, excluding $24$ as a target number. For evaluation, 
we randomly generate 1,000 test cases with distinct target numbers from training set. This setup allows us to test generalization to Game-of-24 when trained on Countdown data. 

\let\origcolorbox\colorbox  %
\renewcommand{\colorbox}[2]{{\setlength{\fboxsep}{0pt}\origcolorbox{#1}{\strut #2}}}

\begin{table}[t]
\vspace{-3em}
\footnotesize
\centering
\caption{\small\textbf{Performance on Countdown and Game-of-24 using Qwen2.5-Math 7B as the base model fine-tuned on Countdown.} 
Each cell shows (success rate / inference time) averaged with 3 seeds, where the time is measured in minutes (m) on an A100 GPU. 
We compare the base LLM with three fine-tuning methods (\colorbox{yellow!30}{\textbf{SFT}}~\citep{ouyang2022training}, \colorbox{green!20}{\textbf{CPO-SimPO}}~\citep{xu2024cpo-simpo}, \colorbox{cyan!20}{\textbf{UFT}} (ours)) under two \textit{learning regimes} (varying learning rate and training data). SimPO~\citep{meng2024simpo} is omitted due to near-zero success rate. 
\textbf{Bolded numbers} indicate the best success rate per row.
\autoref{fig:teaser} visualizes the stronger result per fine-tuning method across \textit{learning regimes}.
}
\vspace{-0.5em}
\setlength{\tabcolsep}{0.25em}
\begin{tabular}{l|c|ccc|ccc}
\toprule
\multirow{2}{*}{\makecell[c]{\textbf{Test set}\\\textbf{\& Inference}}} &
\multirow{2}{*}{\makecell[c]{\textbf{Base LLM}}} &
\multicolumn{3}{c|}{\cellcolor{pink!20}lr=5e-6, CoT+BFS+DFS training data} &
\multicolumn{3}{c}{\cellcolor{magenta!8}lr=1e-6, CoT training data} \\
\cmidrule{3-5}\cmidrule{6-8}
& &
\cellcolor{yellow!30}\textbf{SFT} {\scriptsize($\alpha$=0)} &
\cellcolor{green!20}\textbf{CPO-SimPO} &
\cellcolor{cyan!20}\textbf{UFT} {\scriptsize($\alpha$=1e-3)} &
\cellcolor{yellow!30}\textbf{SFT} {\scriptsize($\alpha$=0)} &
\cellcolor{green!20}\textbf{CPO-SimPO} &
\cellcolor{cyan!20}\textbf{UFT} {\scriptsize($\alpha$=1e-4)} \\
\midrule
\multicolumn{8}{l}{\textbf{Countdown} (1000 cases)}\\
\quad CoT (greedy)  & 6.2\% / 1.0m  & 55.4\% / 0.6m   &  50.6\%  / 0.7m  & \textbf{57.4\%} / 0.5m & 24.7\% / 0.5m & 23.4\% / 0.8m & 24.7\% / 0.5m \\
\quad CoT (pass@1)  & 5.6\% / 3.4m  &  51.8\% / 3.5m  & 45.7\% / 4.7m & \textbf{53.5\%} / 3.4m & 24.1\% / 2.9m & 22.1\% / 5.7m & 24.7\% / 3.3m \\
\quad search: ToT   & 25.0\% / 90m  & 2.4\% / 104m  &     4.1\% / 61m   & 0.0\% / 1.9m          & 22.2\% / 96m  & 23.3\% / 179m & \textbf{28.6\%} / 153m \\
\quad search: RAP   & 10.2\% / 278m &  7.3\% / 155m  &    7.0\% / 137m    & 0.0\% / 341m         & 17.5\% / 141m & 17.7\% / 190m & \textbf{28.9\%} / 439m \\
\midrule
\multicolumn{8}{l}{\textbf{Game-of-24} (100 cases)}\\
\quad CoT (greedy)  & 6.0\% / 0.1m  & 35.0\% / 0.1m &   24.7\% /  0.1m     & \textbf{42.0\%} / 0.1m&  8.0\% / 0.1m &  6.7\% / 0.1m &  8.3\% / 0.1m \\
\quad CoT (pass@1)  & 6.0\% / 0.6m  & 28.3\% / 0.7m  &  23.0\% / 1.3m  & \textbf{29.7\%} / 0.8m&  7.3\% / 0.6m &    7.0\% / 1.5m &  6.7\% / 0.8m \\
\quad search: ToT   & 28.0\% / 6.5m &  2.3\% / 8.2m  &      1.0\% / 4.3m   & 0.3\% / 0.4m         & \textbf{30.7\%} / 8m   & 22.7\% / 13m & 24.0\% / 13m \\
\quad search: RAP   & 27.0\% / 8.3m &  1.3\% / 13m  &  3.0\% / 19m     & 3.7\% / 29m          & 27.7\% / 16m  & 25.7\% / 42m & \textbf{33.7\%} / 75m \\
\bottomrule
\end{tabular}
\vspace{-0.5em}
\label{tab:inference-methods-7b}
\end{table}

\subsection{Setup}
\vspace{-0.5em}

\paragraph{Base LLMs.} We use Qwen2.5-Math 1.5B (Q1.5B) and Qwen2.5-Math 7B (Q7B)~\citep{yang2024qwen2} as base LLMs. Qwen2.5-Math series were state-of-the-art open-weight \textit{mathematical} LLMs  as of December 2024.

\vspace{-0.5em}
\paragraph{Step 1: Data generation setup.} We mainly build upon the LLM-reasoners repository\footnote{\url{https://github.com/maitrix-org/llm-reasoners}}~\citep{hao2024llm}, which provides a unified library for all the LLM reasoners studied here. We significantly accelerate inference by batching inputs using vLLM~\citep{kwon2023efficient}. All reported inference time in our results is based on our optimized codebase.
For each LLM reasoner algorithm, we vary its hyperparameters to gather data from $\mathcal X_{\text{train}}$. 
For CoT~\citep{wei2022chain}, we vary the temperature between $(0.5, 0.7, 1.0)$ and top-p between $(0.7, 0.8, 0.9)$ to have 9 variants. For each variant, we use a fixed 5-shot prompt template and collect $100$ paths for each case in Game-of-24 and $3$ paths for Countdown. For Game-of-24, we follow LLM-reasoners to implement ToT~\citep{yao2023tree} (varying the beam size from $5$ to $16$) and RAP~\citep{hao2023reasoning} (varying the exploration parameter from $1.0$ to $10.0$). 
Since Countdown is implemented in Stream-of-Search (SoS)~\citep{gandhi2024stream}, we use SoS code to implement classic BFS and DFS algorithms, which do not use LLMs but rely on the oracle verifier. We merge the reasoning paths generated by all the variants of the same reasoner, and then remove duplicates by exact matching, to serve as our training dataset. See \autoref{sec:llm_reasoners} for all details.\looseness=-1

\vspace{-0.5em}
\paragraph{Step 2: Fine-tuning setup and baselines.}
We follow the alignment handbook~\citep{Tunstall_The_Alignment_Handbook} to implement all fine-tuning methods on an instance with 8 A100 GPUs. All methods share the same datasets and use common hyperparameters when applicable.
We fix the batch size $B$ to $128$ and train $E=10$ epochs on each dataset for Game-of-24 and $E=2$ epochs for Countdown. We use a cosine schedule and sweep the \textit{peak} learning rate $\eta$ over (1e-5, 5e-6, 2e-6) for Q1.5B and over (5e-6, 2e-6, 1e-6) for Q7B. 
We set $\alpha = 0$ to have the \textbf{SFT} baseline and sweep $\alpha$ over (1e-3, 1e-4, 1e-5, 1e-6) for our UFT. 
We also implement two \textbf{preference-based baselines}: \textbf{SimPO}~\citep{meng2024simpo}, which optimizes over \textit{paired} preference data constructed from successful and failed reasoning paths, and \textbf{CPO-SimPO}~\citep{xu2024cpo-simpo}, which augments SimPO with NLL term to better support reasoning tasks. Both methods are reference-free and leverage negative data as UFT. Notably, CPO-SimPO closely resembles the non-iterative variant of RPO~\citep{pang2024iterative}.
See \autoref{sec:training} for details and \autoref{tab:pref} for method comparison.

\vspace{-0.5em}
\paragraph{Step 3: Evaluation on CoT and search-based reasoning.} Since our training dataset are all CoT-style reasoning paths, fine-tuning on these data should directly improve CoT reasoning capability.   
This is measured by performing \textit{zero-shot} CoT (via greedy decoding or sampling) on $\mathcal X_{\text{test}}$ with fine-tuned LLM. On the other hand, we also aim to retain search-based reasoning capability after fine-tuning. This is measured by performing ToT or RAP on $\mathcal X_{\text{test}}$.\looseness=-1

\vspace{-0.5em}
\paragraph{Inference-time baselines on CoT and search-based reasoning.} Inference-time baselines refer to reasoners that use base LLMs without fine-tuning. 
For CoT reasoning, we use the \textit{few-shot} CoT (via greedy-decoding or sampling) on the base LLMs.
For search-based reasoning, we report the \textit{best variants} of ToT and RAP with the base LLM.
The results for these baselines are shown in \autoref{tab:inference-methods-7b} for Q7B and \autoref{tab:inference-methods-1.5b} for Q1.5B.
We find that search-based reasoning (especially ToT) significantly improves success rates over CoT reasoning for Q7B (6.2\% $\to$ 25\% in Countdown, 6\% $\to$ 28\% in Game-of-24), but not for Q1.5B (2.2\% $\to$ 2.2\% in Countdown, 2\% to 8\% in Game-of-24). However, search-based reasoning substantially increases inference time compared to CoT; for example, in Countdown, ToT and RAP are 90$\times$ and 278$\times$ slower than CoT (greedy decoding).

\newcommand{\agg}[2]{$#1 {\scriptstyle \pm #2}$}

\begin{table}[t]
\vspace{-3em}
\footnotesize
    \centering
    \caption{\small\textbf{CoT reasoning performance of LLMs fine-tuned on Countdown.} Each row shows the averaged result (3 seeds) of an LLM fine-tuned on data from specified sources. Models are evaluated on Countdown (1000 cases) and Game-of-24 (100 cases) test sets using zero-shot CoT via greedy decoding (greedy) or sampling (pass@1). BFS+DFS data are generated by oracle reasoners without LLMs. We highlight \colorbox{cyan!20}{our two contributions} (incorporating search-derived data and the fine-tuning method UFT) and \colorbox{orange!20}{\textbf{the best cell}} for each (test set, CoT inference) pair.
    \textbf{Bold} indicates the highest mean for each row. A superscript $^*$ marks results that are statistically significant according to a Welch’s t-test (p < 0.05).
    \looseness=-1}
    \vspace{-1em}

\begin{subtable}{\linewidth}
\centering
\caption{Base LLM: Qwen2.5-Math 7B (best succ rate with search: 25\% on Countdown, 28\% on Game-of-24)}
\vspace{-0.5em}
\begin{tabular}{l | c c c c}
  \toprule
   & SFT & SimPO & CPO-SimPO & \cellcolor{cyan!20}UFT \\
  \midrule
  \cellcolor{gray!20}\textbf{CoT data} {\scriptsize (quality: \textbf{37.6\%)}} & & \\
  \quad Countdown succ (greedy)   & \agg{32.5\%}{0.3\%} & \agg{0.0\%}{0.0\%} & \agg{27.8\%}{0.5\%} &\agg{\textbf{33.5\%}}{1.0\%} \\
  \quad Countdown succ (pass@1)   & \agg{32.2\%}{0.6\%} & \agg{0.0\%}{0.0\%} & \agg{25.9\%}{1.3\%} & \agg{\textbf{32.5\%}}{1.7\%} \\
  \quad Game-of-24 succ (greedy)  & \agg{16.0\%}{1.0\%} & \agg{0.0\%}{0.0\%} & \agg{11.7\%}{3.1\%} &\agg{\textbf{16.7\%}}{2.9\%} \\
  \quad Game-of-24 succ (pass@1)  & \agg{12.3\%}{2.3\%} & \agg{0.0\%}{0.0\%} & \agg{10.3\%}{4.0\%} &\agg{\textbf{14.7\%}}{4.6\%} \\
  \midrule
  \cellcolor{cyan!20}\textbf{BFS+DFS data} {\scriptsize (quality: \textbf{86.2\%})} & & \\
  \quad Countdown succ (greedy)   & \agg{56.1\%}{0.4\%} & \agg{0.0\%}{0.0\%} & \agg{51.4\%}{0.6\%} &\agg{\textbf{57.1\%$^*$}}{0.3\%} \\
  \quad Countdown succ (pass@1)   & \agg{53.1\%}{2.6\%} & \agg{0.0\%}{0.0\%} & \agg{48.6\%}{1.3\%} &\cellcolor{orange!20}\agg{\textbf{54.4\%}}{1.1\%} \\
  \quad Game-of-24 succ (greedy)  & \agg{37.0\%}{1.0\%} & \agg{0.0\%}{0.0\%} & \agg{31.7\%}{2.5\%} &\agg{\textbf{39.7\%}}{2.5\%} \\
  \quad Game-of-24 succ (pass@1)  & \agg{\textbf{28.3\%}}{4.0\%} & \agg{0.0\%}{0.0\%} & \agg{\textbf{28.3\%}}{4.0\%} &\agg{24.7\%}{5.7\%} \\
  \midrule
  \cellcolor{cyan!20}\textbf{CoT+BFS+DFS data} {\scriptsize (quality: \textbf{86.9\%})} & & \\
  \quad Countdown succ (greedy)   & \agg{55.4\%}{1.9\%} & \agg{0.0\%}{0.0\%} & \agg{50.6\%}{0.6\%} &\cellcolor{orange!20}\agg{\textbf{57.4\%}}{0.4\%} \\
  \quad Countdown succ (pass@1)   & \agg{51.8\%}{2.4\%} & \agg{0.0\%}{0.0\%} & \agg{45.7\%}{0.4\%} &\agg{\textbf{53.5\%}}{0.5\%} \\
  \quad Game-of-24 succ (greedy)  & \agg{35.0\%}{1.0\%} & \agg{0.0\%}{0.0\%} & \agg{24.7\%}{3.1\%} &\cellcolor{orange!20}\agg{\textbf{42.0\%$^*$}}{1.7\%} \\
  \quad Game-of-24 succ (pass@1)  & \agg{28.3\%}{6.7\%} & \agg{0.0\%}{0.0\%} & \agg{23.0\%}{3.0\%} &\cellcolor{orange!20}\agg{\textbf{29.7\%}}{4.0\%} \\
  \bottomrule
\end{tabular}
\vspace{0.5em}
\end{subtable}

\begin{subtable}{\linewidth}
\centering
\caption{Base LLM: Qwen2.5-Math 1.5B (best succ rate with search: 2.2\% on Countdown, 8\% on Game-of-24)}
\vspace{-0.5em}
\begin{tabular}{l | c c c c}
  \toprule
   & SFT & SimPO & CPO-SimPO & \cellcolor{cyan!20}UFT \\
  \midrule
  \cellcolor{gray!20}\textbf{CoT data} {\scriptsize (quality: \textbf{20.0\%})} & & \\
  \quad Countdown succ (greedy)   & \agg{24.2\%}{0.3\%} & \agg{0.6\%}{0.6\%} & \agg{21.6\%}{0.6\%}  & \agg{\textbf{26.1\%$^*$}}{0.6\%} \\
  \quad Countdown succ (pass@1)   & \agg{24.4\%}{0.4\%} & \agg{0.6\%}{0.5\%} & \agg{21.3\%}{0.4\%} &\agg{\textbf{26.1\%$^*$}}{0.4\%} \\
  \quad Game-of-24 succ (greedy)  & \agg{13.3\%}{0.6\%} & \agg{0.0\%}{0.0\%} & \agg{8.7\%}{1.5\%} &\agg{\textbf{14.3\%}}{0.6\%} \\
  \quad Game-of-24 succ (pass@1)  & \agg{9.3\%}{2.5\%}  & \agg{0.0\%}{0.0\%} & \agg{7.3\%}{2.1\%} &\agg{\textbf{14.0\%$^*$}}{2.0\%} \\
  \midrule
  \cellcolor{cyan!20}\textbf{BFS+DFS data} {\scriptsize (quality: \textbf{86.2\%})} & & \\
  \quad Countdown succ (greedy)   & \agg{50.8\%}{0.7\%} & \agg{0.0\%}{0.0\%} & \agg{50.5\%}{1.4\%} &\cellcolor{orange!20}\agg{\textbf{53.7\%$^*$}}{1.2\%} \\
  \quad Countdown succ (pass@1)   & \agg{47.5\%}{0.8\%} & \agg{0.0\%}{0.0\%} & \agg{47.4\%}{1.6\%} & \cellcolor{orange!20}\agg{\textbf{51.2\%$^*$}}{1.1\%} \\
  \quad Game-of-24 succ (greedy)  & \agg{29.7\%}{3.2\%} &  \agg{0.0\%}{0.0\%} & \agg{27.3\%}{3.8\%} & \cellcolor{orange!20}\agg{\textbf{32.0\%}}{4.6\%} \\
  \quad Game-of-24 succ (pass@1)  & \agg{\textbf{23.0\%}}{7.0\%} & \agg{0.0\%}{0.0\%} & \agg{20.3\%}{2.5\%} & \agg{\textbf{23.0\%}}{5.2\%} \\
  \midrule
  \cellcolor{cyan!20}\textbf{CoT+BFS+DFS data} {\scriptsize (quality: \textbf{86.3\%})} & & \\
  \quad Countdown succ (greedy)   & \agg{51.1\%}{0.9\%} &  \agg{0.0\%}{0.0\%} & \agg{48.9\%}{0.8\%} & \agg{\textbf{51.5\%}}{0.7\%} \\
  \quad Countdown succ (pass@1)   & \agg{\textbf{47.1\%}}{1.1\%} & \agg{0.0\%}{0.0\%} & \agg{45.3\%}{1.2\%} &\agg{46.5\%}{1.3\%} \\
  \quad Game-of-24 succ (greedy)  & \agg{\textbf{26.3\%}}{1.5\%} & \agg{0.0\%}{0.0\%} & \agg{24.7\%}{2.9\%} &\agg{26.0\%}{2.6\%} \\
  \quad Game-of-24 succ (pass@1)  & \agg{24.0\%}{2.0\%} & \agg{0.0\%}{0.0\%} & \agg{20.7\%}{2.1\%} &\cellcolor{orange!20}\agg{\textbf{25.0\%}}{5.3\%} \\
  \bottomrule
\end{tabular}
\vspace{0.5em}
\end{subtable}

    \label{tab:cot_reasoning-cd4}
\vspace{-1em}
\end{table}

\begin{table}[t]
\vspace{-3em}
    \footnotesize
    \centering
    \caption{\small\textbf{CoT reasoning performance of LLMs fine-tuned \textit{and} evaluated on Game-of-24.} Each row shows the averaged result (4 seeds) of an LLM fine-tuned on data from specified sources. We highlight \colorbox{cyan!20}{our two contributions} (incorporating search-derived data and the fine-tuning method UFT) and \colorbox{orange!20}{\textbf{the best cell}} for each (test set, CoT inference) pair. 
    \textbf{Bold} indicates the highest mean for each row. A superscript $^*$ marks results that are statistically significant according to a Welch’s t-test (p < 0.05).}\looseness=-1
    \vspace{-1em}

\begin{subtable}{\linewidth}
\centering
\caption{Base LLM: Qwen2.5-Math 7B (best succ rate with search: 28\%)}
\vspace{-0.5em}
\begin{tabular}{l | c c c c}
  \toprule
   & SFT & SimPO & CPO-SimPO & \cellcolor{cyan!20}UFT \\
  \midrule
  \cellcolor{gray!20} \textbf{CoT data} {\scriptsize (quality: \textbf{92.9\%})} & & & & \\
  \quad succ (greedy)   & \agg{28.2\%}{1.7\%} & \agg{2.2\%}{1.3\%} & \agg{\textbf{28.5\%}}{1.9\%} & \agg{28.2\%}{3.6\%} \\
  \quad succ (pass@1)   & \agg{22.5\%}{3.4\%} & \agg{2.2\%}{1.5\%} & \cellcolor{orange!20}\agg{\textbf{27.5\%$^*$}}{2.4\%} & \agg{23.2\%}{3.4\%} \\
  \midrule
  \cellcolor{cyan!20}\textbf{ToT+RAP data} {\scriptsize (quality: \textbf{69.3\%})} & & & & \\
  \quad succ (greedy)   & \agg{13.2\%}{3.6\%} & \agg{0.0\%}{0.0\%} & \agg{\textbf{16.5\%}}{2.9\%} & \agg{15.2\%}{2.2\%} \\
  \quad succ (pass@1)   & \agg{13.2\%}{2.8\%} & \agg{0.0\%}{0.0\%} & \agg{15.0\%}{4.1\%} & \agg{\textbf{17.2\%}}{1.0\%} \\
  \midrule
  \cellcolor{cyan!20}\textbf{CoT+ToT+RAP data} {\scriptsize (quality: \textbf{95.3\%})} & & & & \\
  \quad succ (greedy)   & \agg{27.5\%}{4.2\%} & \agg{0.2\%}{0.5\%} & \agg{26.8\%}{5.7\%} & \cellcolor{orange!20}\agg{\textbf{30.2\%}}{2.1\%} \\
  \quad succ (pass@1)   & \agg{24.2\%}{5.2\%} & \agg{0.2\%}{0.5\%} & \agg{22.8\%}{1.7\%} & \agg{\textbf{26.2\%}}{2.2\%} \\
  \bottomrule
\end{tabular}
\vspace{0.5em}
\end{subtable}

\begin{subtable}{\linewidth}
\centering
\caption{Base LLM: Qwen2.5-Math 1.5B (best succ rate with search: 8\%)}
\vspace{-0.5em}
\begin{tabular}{l | c c c c}
  \toprule
   & SFT & SimPO & CPO-SimPO & \cellcolor{cyan!20}UFT \\
  \midrule
  \cellcolor{gray!20}\textbf{CoT data} {\scriptsize (quality: \textbf{82.1\%})} & & & & \\
  \quad succ (greedy)   & \agg{14.5\%}{3.8\%} & \agg{0.0\%}{0.0\%} & \agg{\textbf{18.5\%}}{1.9\%} & \agg{18.2\%}{3.2\%} \\
  \quad succ (pass@1)   & \agg{17.8\%}{3.4\%} & \agg{0.0\%}{0.0\%} & \agg{17.3\%}{0.5\%} & \agg{\textbf{18.5\%}}{3.5\%} \\
  \midrule
  \cellcolor{cyan!20}\textbf{ToT+RAP data} {\scriptsize (quality: \textbf{44.7\%})} & & & & \\
  \quad succ (greedy)   & \agg{9.5\%}{0.6\%} & \agg{0.0\%}{0.0\%} & \agg{\textbf{10.3\%}}{1.9\%} & \agg{8.0\%}{1.4\%} \\
  \quad succ (pass@1)   & \agg{\textbf{10.0\%}}{1.4\%} & \agg{0.0\%}{0.0\%} & \agg{9.5\%}{1.3\%} & \agg{9.5\%}{2.6\%} \\
  \midrule
  \cellcolor{cyan!20}\textbf{CoT+ToT+RAP data} {\scriptsize (quality: \textbf{87.6\%})} & & & & \\
  \quad succ (greedy)   & \agg{20.0\%}{2.7\%} & \agg{0.0\%}{0.0\%} &  \cellcolor{orange!20}\agg{\textbf{21.8\%}}{1.7\%} & \agg{20.5\%}{3.7\%} \\
  \quad succ (pass@1)   & \agg{17.0\%}{2.2\%} & \agg{0.0\%}{0.0\%} & \agg{18.0\%}{2.6\%} & \cellcolor{orange!20}\agg{\textbf{18.8\%}}{1.3\%} \\
  \bottomrule
\end{tabular}
\vspace{0.5em}
\end{subtable}

    \label{tab:cot_reasoning-game24}
\vspace{-1em}
\end{table}

\subsection{Boosting Chain-of-Thought Reasoning}
\label{sec:cot_reasoning}

We present CoT reasoning results \textit{at inference time} for LLMs fine-tuned on the Countdown (\autoref{tab:cot_reasoning-cd4}) and the Game-of-24 datasets (\autoref{tab:cot_reasoning-game24}). Similar to  \citet{ye2024beyond}, we also evaluate the models trained on Countdown on the Game-of-24 test set, because conceptually Game-of-24 is a subset of Countdown.  
For both tables, we use the highest learning rate (1e-5 for Q1.5B and 5e-6 for Q7B) that yields best performance in CoT reasoning, leaving further discussion on the learning rate for \autoref{sec:search_reasoning}. For the UFT objective, we fix the coefficient $\alpha$ for the same model across different datasets, selected by the average success rate on validation set $\mathcal X_{\text{valid}}$. Our key findings are summarized below.\looseness=-1 

\vspace{-0.5em}
\paragraph{NLL loss is essential for offline fine-tuning on reasoning tasks.} The most salient observation is that SimPO, while successful in alignment tasks using only a preference loss~\citep{meng2024simpo}, yields near-zero success across model sizes and datasets. In contrast, CPO-SimPO~\citep{xu2024cpo-simpo}, which simply adds an NLL term to SimPO, achieves substantially better results. This highlights the critical role of NLL loss, shared by SFT, CPO-SimPO, and UFT, which explicitly encourages models to follow correct reasoning paths. These findings align with prior results in \citet{pang2024iterative, pal2024smaug}.\looseness=-1

\vspace{-0.5em}
\paragraph{Incorporating high-quality training data is decisive for CoT reasoning.}
Second, among methods that use NLL loss, performance is primarily determined by the quality of the training data, outweighing the effect of forgetting objective. We measure \textbf{data quality} by the \textit{best-of-n success rate}: the fraction of \textit{unique} solved cases out of all the training cases.\footnote{Although ToT and RAP greatly outperform CoT as inference-time methods (\autoref{tab:inference-methods-7b}, \autoref{tab:inference-methods-1.5b}) in Game-of-24, they produce lower-quality training data than CoT (\autoref{tab:cot_reasoning-game24}), because CoT's best-of-n (n=100) success rate far exceeds its average (93\% versus 6\% in Q7B, 82\% versus 5\% in Q1.5B). }  
The best performances in \autoref{tab:cot_reasoning-cd4} and \autoref{tab:cot_reasoning-game24} correspond to the datasets with the highest quality (BFS+DFS data in Countdown, CoT data in Game-of-24). In addition, when the two datasets have similar quality (BFS+DFS vs CoT+BFS+DFS in Countdown, CoT vs CoT+ToT+RAP in Game-of-24), their performances are also alike.
Remarkably, fine-tuning on the high-quality classic search (BFS+DFS) data from Countdown yields significantly better performance on Game-of-24 than models fine-tuned \textit{directly} on Game-of-24 CoT data (37\% vs 28.2\% for Q7B and 29.7\% vs 14.5\% for Q1.5B).
This highlights the importance of incorporating high-quality training data, rather than relying solely on CoT data, as is common practice. 

\vspace{-0.5em}
\paragraph{Unpaired learning and forgetting is more robust than paired one.} Lastly, we analyze the effect of forgetting loss to SFT.
UFT achieves the highest mean (ignoring variance) in 26 out of 36 evaluation settings, and is statistically significantly better than baselines in 7 cases. Given the limited number of seeds (3–4), these results may underestimate the true significance, and we expect the evidence to strengthen with additional runs.
In terms of magnitude, UFT yields an average improvement of \textbf{1.3\%} (with a maximum of \textbf{4\%}) in \autoref{tab:cot_reasoning-game24}, and an average of \textbf{1.5\%} (with a maximum of \textbf{7\%}) in \autoref{tab:cot_reasoning-cd4}.

In contrast, CPO-SimPO shows mixed results. While it improves over SFT with an average gain of 1.2\% and maximum of 5\% in \autoref{tab:cot_reasoning-game24}, it underperforms SFT in most scenarios  in \autoref{tab:cot_reasoning-cd4}, with an average \textbf{drop} of 3.5\%. We believe several factors contribute to the contrasting behavior of CPO-SimPO relative to SFT (and UFT) when fine-tuned on Countdown versus Game-of-24.
First, as shown in Appendix \autoref{tab:data}, the \textbf{paired} preference requirement in CPO-SimPO results in the exclusion of 0.1\% to 5.8\% of correct data on Countdown, while no such filtering occurs for Game-of-24. Since correct data determines data quality, this reduction likely contributes to CPO-SimPO’s weaker performance on Countdown.
Second, the exclusion of unpaired failed data, along with the challenge of balancing preference and NLL objectives, may have a greater impact on the more difficult Countdown task.

\subsection{Mitigating Forgetting in Search-based Reasoning}
\label{sec:search_reasoning}

We investigate whether fine-tuned LLMs can retain search-based reasoning abilities (\eg, ToT, RAP), a question that remains underexplored in the absence of continued pretraining. 
Evaluating this retention is analogous to measuring \textit{backward transfer}~\citep{lopez2017gradient} in continual learning, i.e., whether performance on a prior task (search-based reasoning) degrades after fine-tuning on a new one (CoT-style data). 
We focus on Q7B, which exhibits strong search-based reasoning performance and a clear gap over CoT baselines (\autoref{tab:inference-methods-7b}). For completeness, the results of Q1.5B is shown in \autoref{app:results}.\looseness=-1

Although learning rate decay is employed following the alignment handbook, we find that using a peak learning rate of 2e-5, as in the original Qwen2.5-Math setup~\citep{yang2024qwen2}, leads to \textit{catastrophic forgetting}~\citep{mccloskey1989catastrophic}, with near-zero performance in inference-time search. We hypothesize that this degradation stems from the lack of reward modeling in CoT-style supervision, which would be essential for search-based reasoning (see \autoref{sec:explanation} for details).
\looseness=-1 

\begin{figure}[t]
\vspace{-3em}
    \centering
    \includegraphics[width=0.85\linewidth]{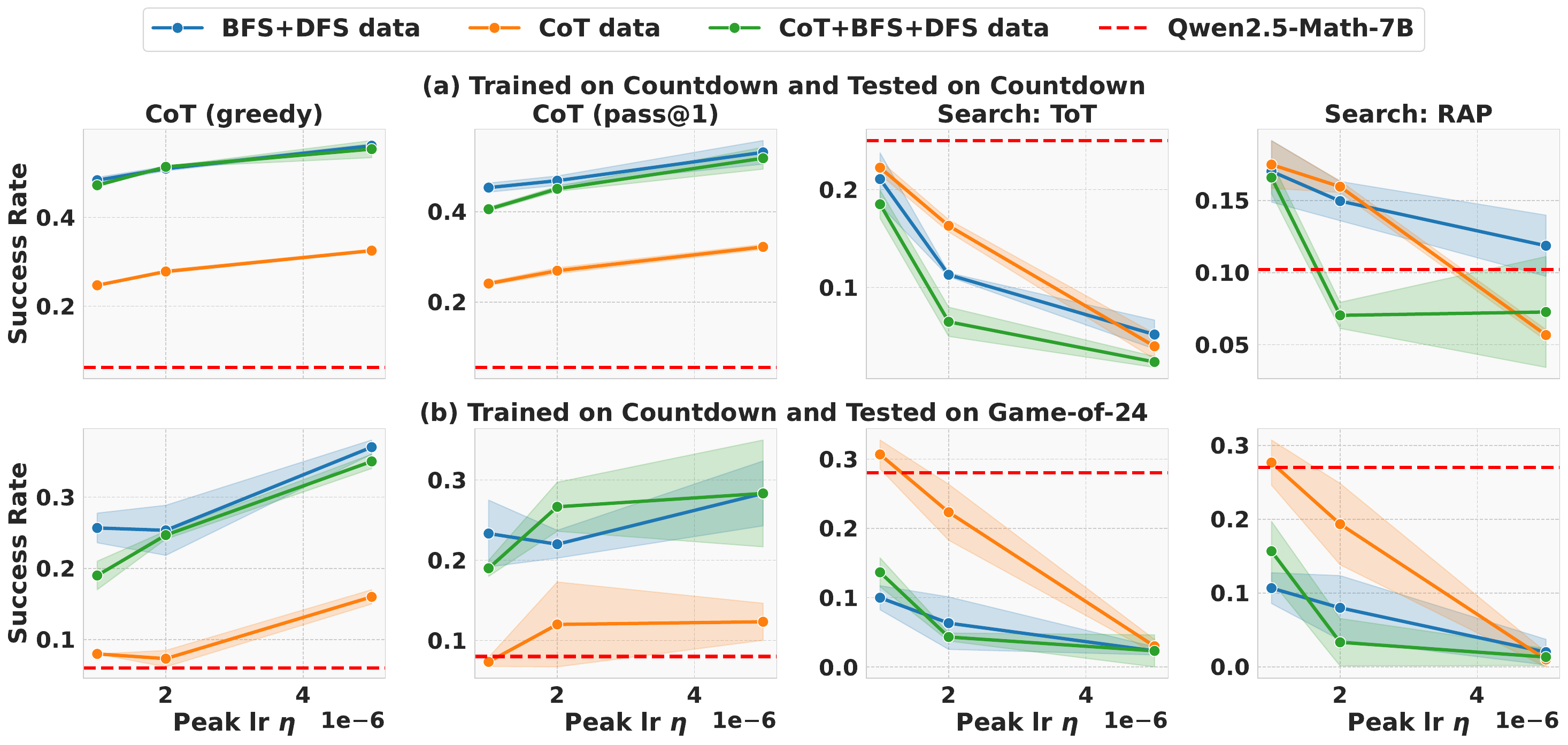}
\vspace{-0.5em}
    \caption{\small\textbf{Impact of (peak) learning rate and data sources on CoT vs. search-based reasoning} for standard SFT ($\alpha$=0) using Qwen2.5-Math 7B as the base model. 
    Learning rate mediates a trade-off between CoT and search performance; CoT data better preserves search capability in most cases.
    }
\vspace{-1em}
    \label{fig:lr_sft-cd4}
\end{figure}

\vspace{-0.5em}
\paragraph{A very low (peak) learning rate is crucial to retaining search capability. }
Inspired by continual learning~\citep{mirzadeh2020understanding}, we adopt a much smaller learning rate to keep parameters within the ``basic capability basin''~\citep{chen2025understanding}, preserving pretrained knowledge. 
\autoref{fig:lr_sft-cd4} and \autoref{fig:lr_sft-g24} shows the effects of (peak) learning rate in standard SFT on Q7B. As we reduce the learning rate from 5e-6 to 1e-6 (an exceptionally low value for SFT), we observe a clear trade-off: CoT reasoning performance declines rapidly (although it remains above the baseline), while inference-time search improves significantly (although often remains below the baseline). This behavior is similar to learning rate effects seen in reference-free preference optimization~\citep{meng2024simpo}.

\vspace{-0.5em}
\paragraph{CoT training data is more effective for preserving search capability.} 
An interesting observation from \autoref{fig:lr_sft-cd4} to \autoref{fig:lr_sft-g24} is that in both Countdown and Game-of-24 training settings, CoT data leads to better inference-time search than search-derived data, although CoT data has \textit{much lower} quality compared to BFS+DFS data. 
Both CoT and search-derived data fall under CoT-style paths, but the latter is generated by a search algorithm rather than directly by the LLM. This suggests that CoT data, being more aligned with the LLM’s on-policy distribution, induces less distribution shift during SFT, thereby reducing forgetting.

\begin{figure}[h]
    \centering
    \includegraphics[width=0.85\linewidth]{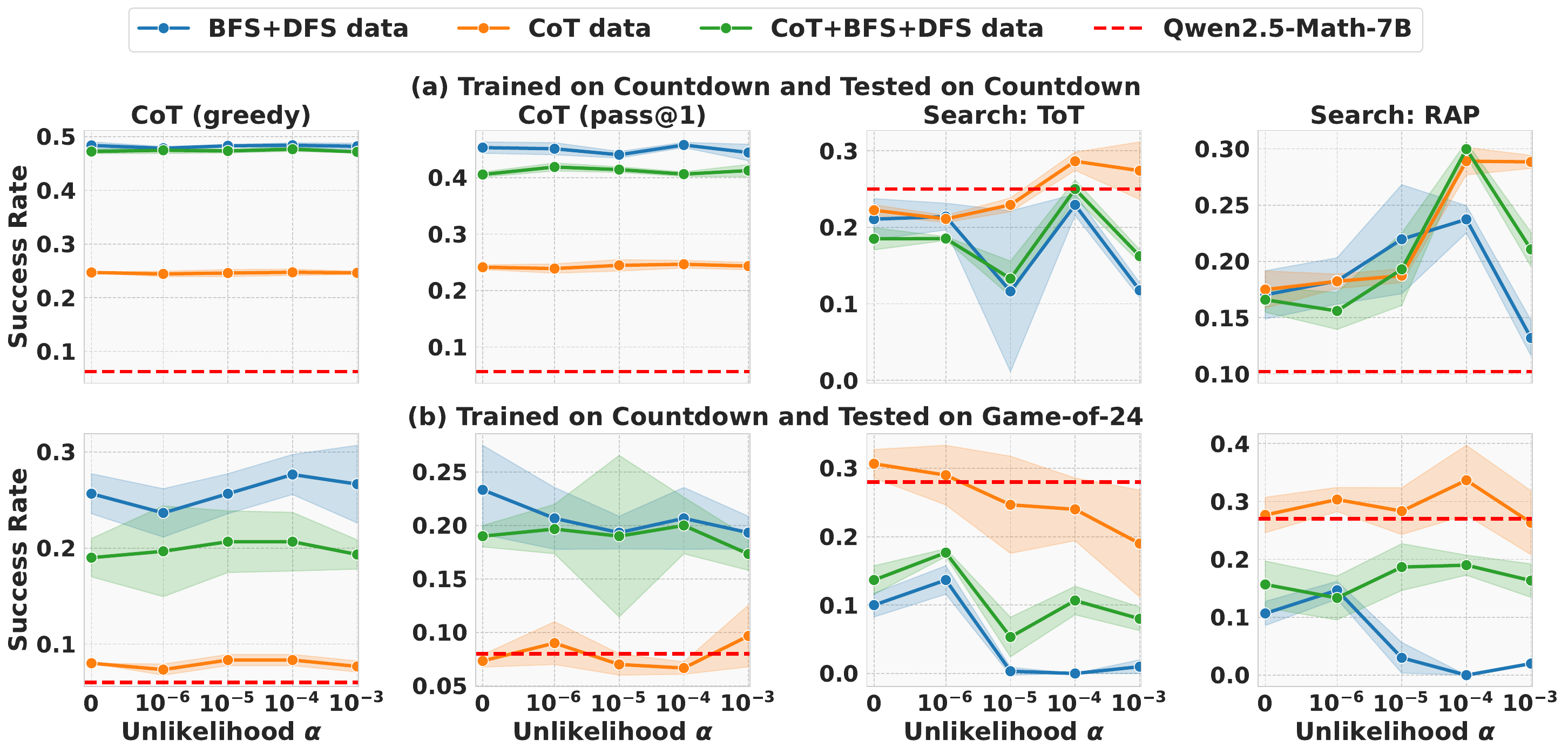}
    \vspace{-0.5em}
    \caption{\small\textbf{Effects of unlikelihood loss and data sources on CoT vs. search-based reasoning} using Qwen2.5-Math 7B as the base model, with the peak learning rate as 1e-6. Unlikelihood loss leads to greater improvements in search performance than in CoT inference in most cases.}
    \label{fig:cft}
    \vspace{-1.5em}
\end{figure}

\vspace{-0.5em}
\paragraph{Unlikelihood auxiliary loss can significantly improve search capability.} Building on the findings in \autoref{fig:lr_sft-cd4}, we observe that the effect of a low learning rate extends to UFT and CPO-SimPO as well. Here, we examine the impact of the unlikelihood loss specifically at the lowest learning rate of 1e-6. Consistent with \autoref{sec:cot_reasoning}, UFT provides marginal improvements over SFT in CoT reasoning, regardless of the choice of $\alpha$. However, with $\alpha$=1e-4, inference-time search improves considerably compared to SFT ($\alpha$=0) in most cases. For example, with the same CoT data, ToT achieves 28.6\% vs. 22.2\% under SFT, and RAP reaches 28.9\% vs. 17.5\% -- far exceeding the RAP baseline of 10.2\%. This indicates that learning to forget incorrect CoT-style paths could generalize to forgetting incorrect intermediate search steps.

\section{Conclusion and Future Work}
We propose a simple and effective method for improving reasoning on arithmetic puzzle domains, addressing three underexplored challenges: (1) augmenting training data by unifying mixed data formats from diverse reasoners into a CoT-style format, (2) leveraging unpaired correct and failed paths via unlikelihood fine-tuning, and (3) mitigating loss of search capability using a small learning rate.
Experiments on Countdown and Game-of-24 show that data quality, unlikelihood loss, and learning rate are key to balancing CoT efficiency and search capability.
Future work can explore more advanced unlearning objectives~\citep{tamirisa2024tamper,huang2024booster}, incorporate stabilization techniques~\citep{meng2024simpo}, and extend the framework to other reasoning-oriented LLMs.

%% file: appendix.tex
\newpage

\section{Fine-tuning Details}
\label{sec:training}

We follow the alignment handbook\footnote{\url{https://github.com/huggingface/alignment-handbook}}~\citep{Tunstall_The_Alignment_Handbook} to fine-tune two base LLMs with full parameters, 
Qwen2.5-Math 1.5B (Q1.5B)\footnote{\url{https://huggingface.co/Qwen/Qwen2.5-Math-1.5B}} and Qwen2.5-Math 7B (Q7B)\footnote{\url{https://huggingface.co/Qwen/Qwen2.5-Math-7B}}~\citep{yang2024qwen2}. Below are common setups for all the methods (i.e., SFT, UFT, and preference-based methods).

We disable sequence packing to prevent cross-contamination\footnote{\url{https://github.com/huggingface/transformers/issues/25452}} and set a context length of 256 tokens. which is sufficient for our tasks. Training is performed in \texttt{bfloat16} precision, and model checkpoints are saved every 5\% epochs for later evaluation.

The learning rate follows \texttt{cosine\_with\_min\_lr} schedule\footnote{\url{https://github.com/huggingface/transformers/blob/a22a4378d97d06b7a1d9abad6e0086d30fdea199/src/transformers/optimization.py\#L338C5-L338C48}}: it linearly warms up from 0 to $\eta$ over the first 10\% epochs, then decays to a minimum of 7e-8 via cosine function for the rest epochs. In \autoref{sec:cot_reasoning}, we use a peak learning rate $\eta$ of 1e-5 for Q1.5B and 5e-6 for Q7B, while in \autoref{sec:search_reasoning}, we decrease $\eta$ to much lower values, as discussed in the main paper. 

\subsection{SFT Baseline and UFT}

Following \autoref{sec:uft}, in our implementation, Unlikelihood Fine-Tuning (UFT) and Supervised Fine-Tuning (SFT) differ only by the unlikelihood loss coefficient $\alpha$. 
A batch size $B = 128$ is used for both successful and failed data batches (after gradient accumulation) in UFT.

\subsection{Preference-based (Reference-free) Baselines: SimPO and CPO-SimPO}

\paragraph{Background.} In addition to SFT, we compare UFT against offline preference-based methods: \textbf{SimPO}~\citep{meng2024simpo} and \textbf{CPO-SimPO}~\citep{xu2024cpo-simpo}, both designed to align LLMs with human preferences. 
While UFT is not a preference-based method per se, our reasoning task can be interpreted through the lens of preference alignment, where successful reasoning paths are preferred over failed ones. Like preference-based approaches, UFT learns from both successful and failed paths, making these methods natural baselines for comparison.
The implementations of these methods follow the TRL library\footnote{\url{https://huggingface.co/docs/trl/en/cpo_trainer}}~\citep{vonwerra2022trl}.

SimPO and CPO-SimPO build on DPO~\citep{rafailov2023direct} and CPO~\citep{xu2024contrastive}, requiring \textit{preference pairs} but eliminating the need for a reference model used in KL regularization (i.e., reference-free). This results in improved training efficiency and reduced GPU memory usage. 
By contrast, while KTO~\citep{ethayarajh2024kto} like UFT operates on \textit{unpaired} data, it still requires a reference model during training. In our preliminary experiments, running KTO using the same global batch size as UFT necessitated precomputing reference model logits and offloading the optimizer to the CPU, resulting in a $3\times$ increase in training time on the Game-of-24 dataset and out-of-memory errors on the Countdown dataset. Consequently, we exclude KTO from our baselines and instead focus on reference-free approaches such as SimPO.

\autoref{tab:pref} summarizes the comparison between UFT and several offline preference-based methods. Notably, UFT is both reference-free and does not require preference pairs. 

\begin{table}[h]
    \caption{\textbf{Brief comparison of selected offline preference-based methods and UFT.}}
    \label{tab:pref}
    \vspace{-0.5em}
    \small
    \centering
    \begin{tabular}{ccc}
       Offline fine-tuning method  & Preference data  & Reference-free?  \\
       \midrule
       DPO~\citep{rafailov2023direct}, (non-iterative) RPO~\citep{pang2024iterative}  & Paired & No \\
       CPO~\citep{xu2024contrastive}, SimPO~\citep{meng2024simpo}, CPO-SimPO~\citep{xu2024cpo-simpo} & Paired & Yes \\
       KTO~\citep{ethayarajh2024kto} & Unpaired & No \\
       UFT (Ours) & Unpaired & Yes  
    \end{tabular}
\end{table}

\paragraph{Simple Preference Optimization
with a Reference-Free Reward (SimPO)~\citep{meng2024simpo}.} SimPO relies on \textbf{paired} preference data, denoted as $\D_{\text{paired}} = \{(\x, \y^+, \y^-)\}$, where $(\y^+,\y^-)$ are successful and failed reasoning paths for the same input $\x$, respectively. To construct $\D_{\text{paired}}$ from our original dataset $\D$,  we first filter for inputs that have both successful and failed paths. For each successful path $(\x, \y^+)$, we generate $E$ preference pairs by randomly sampling $E$ failed paths $(\x, \y^-)$ corresponding to the same input, treating them as rejected responses. This sampling strategy ensures that the resulting dataset is similar in size to that used for UFT and captures as much diversity as possible within the constraint of pairing.

The training objective is defined as:
\begin{equation}
\min_\theta J_{\text{SimPO}}(\theta;\D_{\text{paired}}) =  -\E{(\x, \y^+,\y^-)\sim \D_{\text{paired}}}{\log \sigma\left(\frac{\beta}{|\y^+|} \log \pi_\theta(\y^+ \mid \x) - \frac{\beta}{|\y^-|} \log \pi_\theta(\y^- \mid \x) - \gamma \right)},
\end{equation}
where $\sigma$ is the sigmoid function, $\beta = 0.1$ is a scaling factor, and $\gamma = 0.5$ is the target reward margin, following default values in TRL. The length normalization terms ($|\y^+|$, $|\y^-|$) help reduce bias toward longer responses. This normalization could also be explored in UFT as future work. The margin $\gamma$ enforces a minimum separation between the preferred and rejected log-probabilities.

\paragraph{CPO-SimPO~\citep{xu2024cpo-simpo}.} 
Although \textit{pure} preference-based methods (e.g., DPO, SimPO) perform well in alignment tasks, they have been found to underperform on reasoning tasks that require explicit maximization of preferred responses~\citep{meng2024simpo,pal2024smaug,pang2024iterative}. While CPO-SimPO was not originally motivated by reasoning tasks, we adopt it because CPO-SimPO addresses this issue by augmenting SimPO with the NLL term from CPO~\citep{xu2024contrastive}: 
\begin{equation}
\begin{split}
&\min_\theta J_{\text{CPO-SimPO}}(\theta;\D_{\text{paired}})   \\
&=-\E{(\x, \y^+,\y^-)\sim \D_{\text{paired}}}{\log \sigma\left(\frac{\beta}{|\y^+|} \log \pi_\theta(\y^+ \mid \x) - \frac{\beta}{|\y^-|} \log \pi_\theta(\y^- \mid \x) - \gamma \right) + \lambda \log \pi_\theta(\y^+ \mid \x)},
\end{split}
\end{equation}
where the NLL coefficient $\lambda = 1.0$ following the TRL.

\paragraph{Connection with RPO~\citep{pang2024iterative}.} CPO-SimPO~\citep{xu2024cpo-simpo} resembles a non-iterative version of Iterative Reasoning Preference Optimization (IRPO)~\citep{pang2024iterative}, which we call RPO: 
\begin{equation}
\begin{split}
&\min_\theta J_{\text{RPO}}(\theta;\D_{\text{paired}})   \\
&=-\E{(\x, \y^+,\y^-)\sim \D_{\text{paired}}}{\log \sigma\left(\beta \log \frac{\pi_\theta(\y^+ \mid \x)}{\pi_{\text{ref}}(\y^+ \mid \x)} - \beta\log \frac{\pi_\theta(\y^- \mid \x)}{ \pi_{\text{ref}}(\y^- \mid \x)} \right) + \frac{\lambda}{|\y^+|} \log \pi_\theta(\y^+ \mid \x)},
\end{split}
\end{equation}
where $\pi_{\text{ref}}$ is the reference model (i.e., the base LLM). 
CPO-SimPO becomes equivalent to RPO when a reference model is introduced, the length normalization and reward margin in the preference term are removed, and length normalization is applied in the NLL term instead. However, since RPO, like DPO and KTO, depends on a reference model, we exclude it from our baseline comparisons.

\section{Additional Results on Countdown and Game-of-24}
\label{app:results}

\subsection{Training Dataset Statistics}

\begin{table}[t]
\vspace{-3em}
\small\centering
    \caption{Statistics of training datasets: \textbf{total} correct paths are used by SFT and UFT, while \textbf{paired} correct paths are used by SimPO and CPO-SimPO (shown as a percentage of total). Data quality refers to the best-of-n success rate reported in \autoref{tab:cot_reasoning-cd4} and \autoref{tab:cot_reasoning-game24}.}
\vspace{-0.5em}
    \begin{tabular}{llll}
    \toprule
      Training data (task, base, reasoner)   &  $\#$ Total correct paths & $\#$ Paired correct paths & Quality\\
      \midrule
      Countdown, Q7B, CoT   & 428.0k & 97.50\% & 37.6\% \\ 
      Countdown, N/A, BFS+DFS   & 614.2k & 97.55\% & 86.2\%  \\ 
      Countdown, Q7B, CoT+BFS+DFS   & 985.3k & 99.88\%  & 86.9\% \\
      Countdown, Q1.5B, CoT   & 147.9k & 94.22\%  & 20.0\% \\ 
      Countdown, Q1.5B, CoT+BFS+DFS   & 734.3k & 99.86\% & 86.3\%  \\
      \midrule
      Game-of-24, Q7B, CoT   & 9.1k & 100\%  & 92.9\% \\ 
      Game-of-24, Q7B, ToT+RAP   & 5.1k & 100\% & 69.3\% \\ 
      Game-of-24, Q7B, CoT+ToT+RAP   & 13.7k & 100\% & 95.3\%  \\
      Game-of-24, Q1.5B, CoT   & 4.4k & 100\%  & 82.1\%  \\ 
      Game-of-24, Q1.5B, ToT+RAP   & 1.7k & 100\% & 44.7\%  \\ 
      Game-of-24, Q1.5B, CoT+ToT+RAP   & 6.0k & 100\% & 87.6\%  \\
      \bottomrule
    \end{tabular}
    \label{tab:data}
\end{table}

For Countdown, we train for 2 epochs due to the large dataset size, shown as in \autoref{tab:data}. On an instance with 8×A100 GPUs (40GB), total training time on CoT+BFS+DFS data is under 5 hours for Q1.5B and under 30 hours for Q7B using UFT.
For Game-of-24, we train for 10 epochs. Total training time is under 0.5 hours for Q1.5B and under 2.5 hours for Q7B using UFT.
SimPO and CPO-SimPO only require a single dataloader over paired data, resulting in roughly 2× shorter training time compared to UFT.

We emphasize that our goal is to reduce \textbf{inference time} in reasoning. While training may be relatively expensive, it is performed offline and does not impact deployment efficiency or user-facing latency.

\subsection{Additional Results with SFT and UFT}

\begin{figure}[H]
    \centering
    \includegraphics[width=0.85\linewidth]{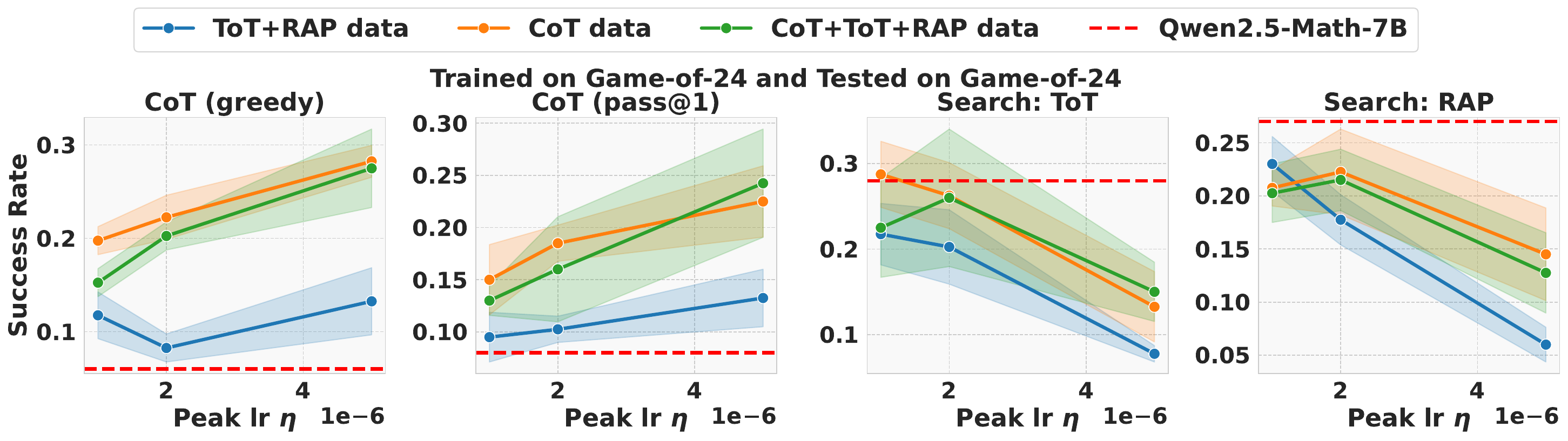}
    \vspace{-0.5em}
    \caption{Results of standard SFT ($\alpha$=0) using \textbf{Qwen2.5-Math 7B} as the base model for \textbf{Game-of-24}. The same conclusion in \autoref{fig:lr_sft-cd4} holds.}
    \vspace{-0.5em}
    \label{fig:lr_sft-g24}
\end{figure}

\begin{figure}[H]
\vspace{-3em}
    \centering
    \includegraphics[width=0.85\linewidth]{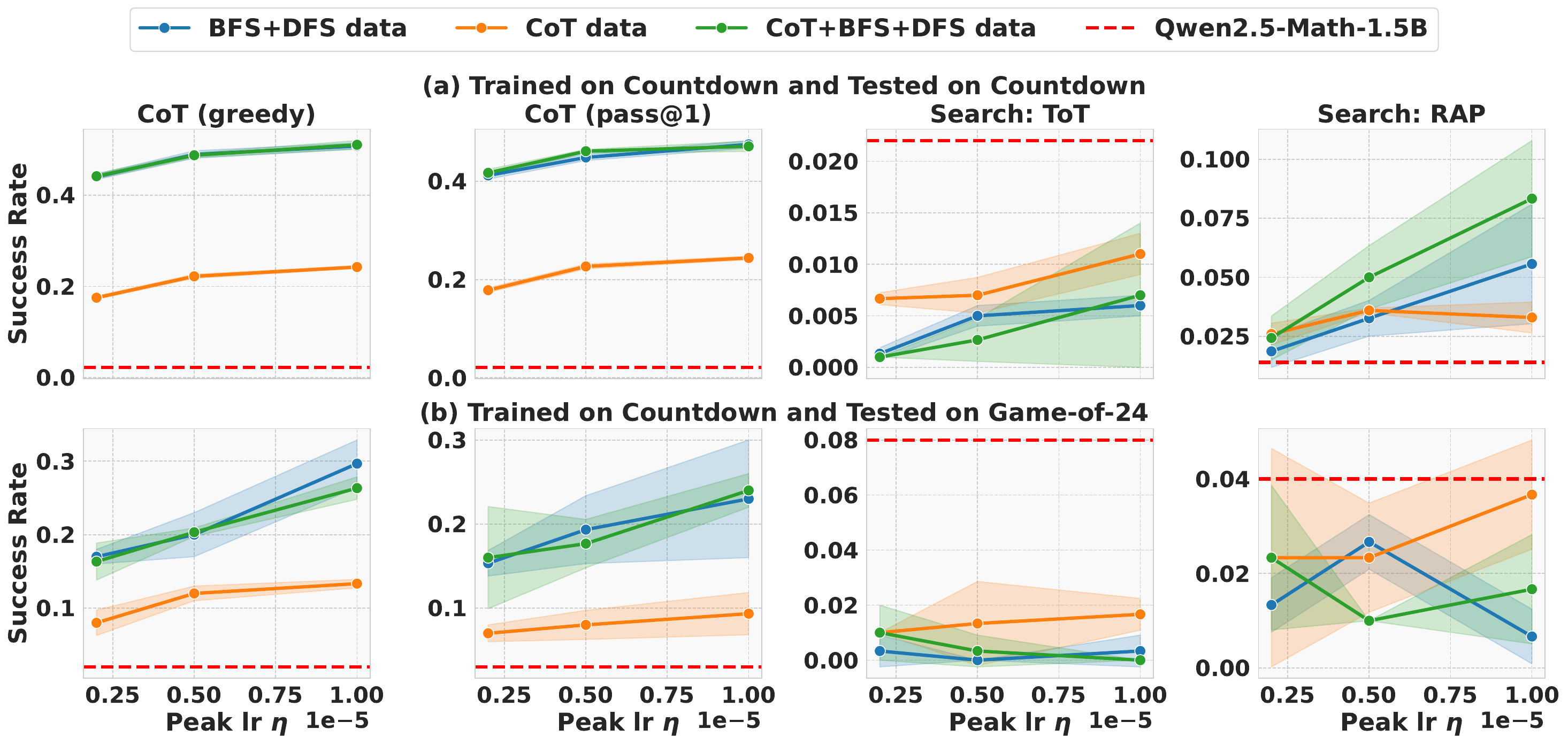}
    \includegraphics[width=0.85\linewidth]{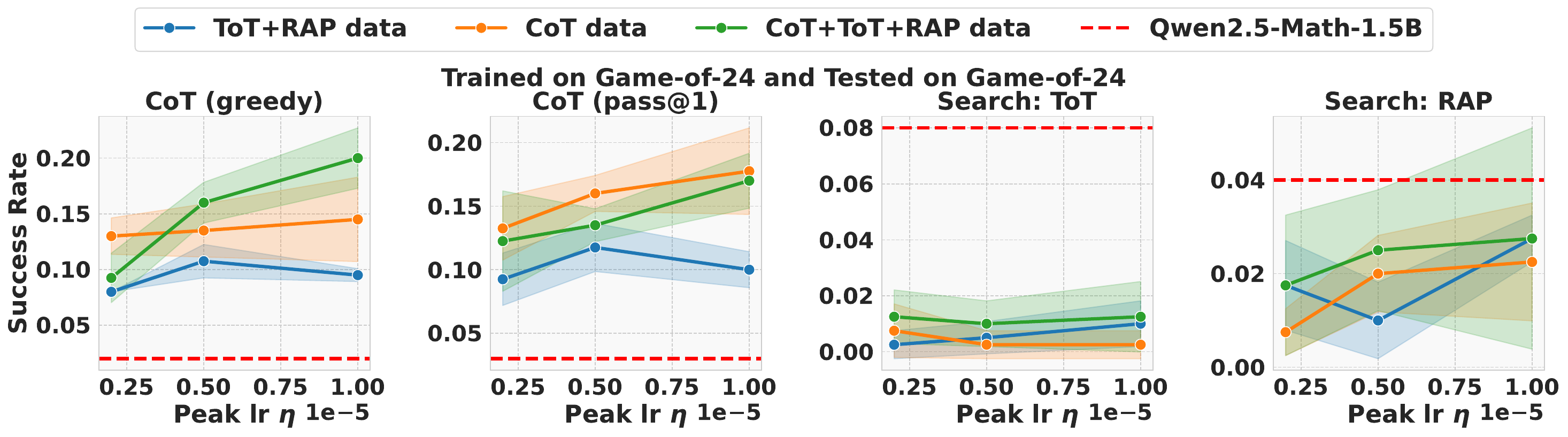}
    \vspace{-0.5em}
    \caption{Results of standard SFT ($\alpha$=0) using \textbf{Qwen2.5-Math 1.5B} as the base model for \textbf{Countdown} (top) and \textbf{Game-of-24} (bottom). Due to the base model's weak search ability, learning rate and data source have little observable effect.}
    \vspace{-0.5em}
    \label{fig:lr_sft-1.5b}
\end{figure}

\begin{figure}[H]
    \centering
    \includegraphics[width=0.85\linewidth]{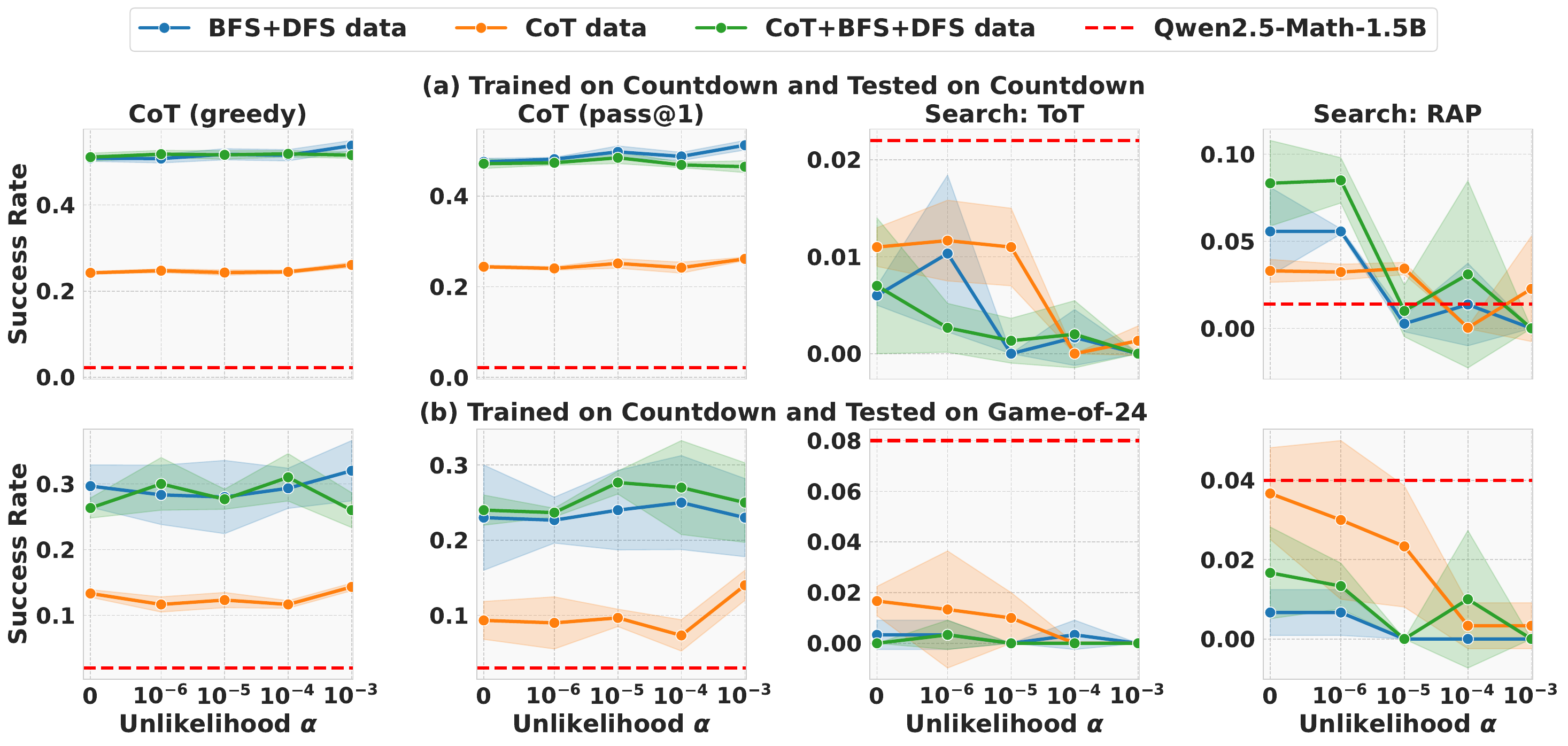}
    \vspace{-0.5em}
    \caption{Results of UFT using \textbf{Qwen2.5-Math 1.5B} as the base model for \textbf{Countdown}, with the peak learning rate as 1e-5. Due to the base model's weak search ability, unlikelihood loss has little observable effect.}
    \vspace{-0.5em}
    \label{fig:cft-1.5b}
\end{figure}

For completeness, we provide fine-tuning results based on Qwen2.5-Math 1.5B in \autoref{fig:lr_sft-1.5b} and \autoref{fig:cft-1.5b}.  Given its weak search-based reasoning performance as a base model (e.g., success rate $\approx$2\% in Countdown) shown in \autoref{tab:inference-methods-1.5b}, the effects of learning rate, CoT data, and unlikelihood loss are minimal, likely due to its poor initial capability. This suggests that the observed benefits depend on the model capability.

\begin{table}[H]
\small
\centering
\caption{Performance on Countdown and Game-of-24 test sets, using \textbf{Qwen2.5-Math 1.5B} as the base model. Due to its weak search capability, we omit fine-tuning results here (please see \autoref{tab:cot_reasoning-cd4} and \autoref{tab:cot_reasoning-game24} for fine-tuning results on CoT inference).}
\vspace{-0.5em}
\begin{tabular}{l|c}
\toprule
\makecell[c]{\textbf{Test set}\\\textbf{\& Inference}} &
\makecell[c]{\textbf{Base LLM}} \\
\midrule
\multicolumn{2}{l}{\textbf{Countdown} (1000 cases)}\\
\quad CoT (greedy)  & 2.2\% / 0.6m   \\
\quad CoT (pass@1)  & 2.1\% / 3.2m   \\
\quad search: ToT   & 2.2\% / 68m   \\
\quad search: RAP   & 1.4\% / 143m  \\
\midrule
\multicolumn{2}{l}{\textbf{Game-of-24} (100 cases)}\\
\quad CoT (greedy)  & 2.0\% / $<$0.1m \\
\quad CoT (pass@1)  & 3.0\% / 0.5m   \\
\quad search: ToT   & 8.0\% / 5.4m   \\
\quad search: RAP   & 4.0\% / 11.3m  \\
\bottomrule
\end{tabular}
    \label{tab:inference-methods-1.5b}
\end{table}

\subsection{Additional Results with Alternative Unlearning Methods}
\label{app:unlearning}

\paragraph{Gradient Ascent and RepNoise.} To further evaluate our framework, we conduct additional experiments by replacing the unlikelihood (UL) loss $J_{\text{UL}}$ in \autoref{eq:obj} with other established unlearning objectives, while keeping all other settings identical. In particular, we examine Gradient Ascent (GA)~\citep{yao2023large} and RepNoise~\citep{rosati2024representation}.

GA is a simple baseline for unlearning that aims to minimize the probability of failed paths: 

\begin{equation}
J_{\text{GA}}(\theta;\D^-) \defeq \E{(\x,\y^-)\sim \D^-}{\log \pi_\theta(\y^- \mid \x)}.
\end{equation}
Although GA is shown to be unstable during optimization~\citep{zhang2024negative}, we mitigate this issue by using a small loss coefficient of $\alpha = $ 1e-5, which is on a similar magnitude as that used for UL. This choice substantially improves numerical stability in our experiments.

Representation Noising (RepNoise)~\citep{rosati2024representation}, on the other hand, is a recent advanced unlearning method designed for enhancing LLM safety. It combines Gradient Ascent with removing harmful representations,  encouraging the model’s representations (per-layer hidden states) for negative inputs to align with random noise. Within our framework, its objective is written as: 
\begin{equation}
\begin{split}
&J_{\text{RepNoise}}(\theta;\D^-) \\
& \defeq \log\left(\frac{1}{L}\E{(\x,\y^-)\sim \D^-}{\sum_{l=1}^L \log \pi_{\theta_{\text{head}}}(\y^- \mid h_l^{\text{post}}(\x))}\right) + \frac{\beta}{L} \E{\x\sim \D^-}{\sum_{l=1}^L\mathrm{MMD}(h_l^{\text{pre}}(\x), \mathcal N(0,I))}.
\end{split}
\end{equation}
The first term acts as a generalized gradient-ascent objective (with an additional log-transformation), reducing the likelihood of undesired outputs based on the per-layer post-residual hidden states $h_l^{\text{post}}(\x)$ passed into the model’s output head $\pi_{\theta_{\text{head}}}$, where $L=28$ denotes the total number of layers. The second term enforces that the pre-residual hidden states $h_l^{\text{pre}}(\x)$ to be close to Gaussian noise according to the maximum mean discrepancy (MMD). Following the original implementation, we set the RepNoise loss coefficient to $\alpha = 0.5$ and its coefficient for the MMD term $\beta=$ 1e-3.  

\paragraph{CoT reasoning results for Gradient Ascent and RepNoise.} \autoref{tab:cot_reasoning-unlearning} shows results for CoT reasoning with the same learning rate for all unlearning methods on Game-of-24. GA achieves performance comparable to SFT and UFT, with slightly lower averages than UFT. RepNoise, which places strong emphasis on the unlearning objective (via equal weighting and representation-level forgetting), performs consistently worse than the other methods. These observations suggest that directly applying unlearning methods originally designed for LLM safety may require additional design considerations.

Across all unlearning methods, our results consistently support the claim that higher training data quality, regardless of whether the data originate from CoT or search-based algorithms, leads to better fine-tuning performance.

\paragraph{Search-based reasoning results for Gradient Ascent and RepNoise.}  \autoref{tab:unlearning_methods_lr} further examines the impact of learning rate and dataset source on CoT and search-based reasoning. Following the previous setup, we sweep learning rates over $\{$1e-6, 2e-6, 5e-6$\}$ for the 7B base model. 

For GA, our key findings for SFT and UFT (\autoref{sec:search_reasoning}) remain valid. A smaller learning rate slightly reduces CoT reasoning performance but can substantially preserve, or even improve, the base model’s search capability (e.g., 31\% in ToT and 28\% in RAP). Moreover, CoT training is more effective for maintaining search ability compared to mixed data (CoT+ToT+RAP), suggesting that using more on-policy data reduces distribution shift and mitigates forgetting, as discussed in \autoref{sec:search_reasoning}.

In contrast, RepNoise exhibits a drastic drop in search-based performance (ToT and RAP), reaching 0.0\% regardless of the learning rate, even though its CoT reasoning improves over the base LLM. We hypothesize that this degradation arises from the large default coefficient ($\alpha = 0.5$) in the RepNoise loss, which amplifies the effective learning rate for unlearning far beyond those used in other methods such as UFT and GA ($\leq 10^{-3}$). This aggressive unlearning, coupled with representation-level regularization toward random noise, likely disrupts internal representations crucial for search behaviors (e.g., hypothesis branching and reward estimation), even when the training objective (CoT reasoning) itself remains stable. These results underscore the importance of carefully tuning the learning rate and loss coefficient when extending unlearning objectives to LLM reasoning domains.

\begin{table}[t]
    \footnotesize
    \centering
    \caption{\small\textbf{Additional results with alternative unlearning methods: CoT reasoning performance of LLMs fine-tuned and evaluated on Game-of-24.} Each cell shows the averaged result (4 seeds) of an LLM fine-tuned on data from specified sources. We highlight \colorbox{cyan!20}{our two contributions} (incorporating search-derived data and the fine-tuning method UFT). \looseness=-1}
    \vspace{-1em}

\begin{subtable}{\linewidth}
\centering
\caption{Base LLM: Qwen2.5-Math 7B (best succ rate with search: 28\%)}
\vspace{-0.5em}
\begin{tabular}{l | c c c c}
  \toprule
   & SFT & \cellcolor{cyan!20}UFT & GA & RepNoise \\
  \midrule
  \cellcolor{gray!20} \textbf{CoT data} {\scriptsize (quality: \textbf{92.9\%})} & & & & \\
  \quad succ (greedy)   & \agg{28.2\%}{1.7\%} & \agg{28.2\%}{3.6\%} & \agg{26.8\%}{2.1\%} & \agg{20.5\%}{3.7\%} \\
  \quad succ (pass@1)   & \agg{22.5\%}{3.4\%} & \agg{23.2\%}{3.4\%} & \agg{25.5\%}{2.4\%} & \agg{17.2\%}{1.0\%} \\
  \midrule
  \cellcolor{cyan!20}\textbf{ToT+RAP data} {\scriptsize (quality: \textbf{69.3\%})} & & & & \\
  \quad succ (greedy)   & \agg{13.2\%}{3.6\%} & \agg{15.2\%}{2.2\%} & \agg{13.5\%}{5.1\%} & \agg{12.8\%}{3.0\%} \\
  \quad succ (pass@1)   & \agg{13.2\%}{2.8\%} & \agg{17.2\%}{1.0\%} & \agg{15.8\%}{2.1\%} & \agg{13.2\%}{5.1\%} \\
  \midrule
  \cellcolor{cyan!20}\textbf{CoT+ToT+RAP data} {\scriptsize (quality: \textbf{95.3\%})} & & & & \\
  \quad succ (greedy)   & \agg{27.5\%}{4.2\%} & \agg{30.2\%}{2.1\%} & \agg{28.5\%}{1.7\%} & \agg{20.8\%}{2.1\%} \\
  \quad succ (pass@1)   & \agg{24.2\%}{5.2\%} & \agg{26.2\%}{2.2\%} & \agg{25.5\%}{3.1\%} & \agg{19.0\%}{2.2\%} \\
  \bottomrule
\end{tabular}
\vspace{0.5em}
\end{subtable}

\begin{subtable}{\linewidth}
\centering
\caption{Base LLM: Qwen2.5-Math 1.5B (best succ rate with search: 8\%)}
\vspace{-0.5em}
\begin{tabular}{l | c c c c}
  \toprule
   & SFT & \cellcolor{cyan!20}UFT & GA & RepNoise \\
  \midrule
  \cellcolor{gray!20}\textbf{CoT data} {\scriptsize (quality: \textbf{82.1\%})} & & & & \\
  \quad succ (greedy)   & \agg{14.5\%}{3.8\%} & \agg{18.2\%}{3.2\%} & \agg{15.8\%}{2.8\%} & \agg{12.2\%}{3.3\%} \\
  \quad succ (pass@1)   & \agg{17.8\%}{3.4\%} & \agg{18.5\%}{3.5\%} & \agg{17.0\%}{2.2\%} & \agg{12.0\%}{2.2\%} \\
  \midrule
  \cellcolor{cyan!20}\textbf{ToT+RAP data} {\scriptsize (quality: \textbf{44.7\%})} & & & & \\
  \quad succ (greedy)   & \agg{9.5\%}{0.6\%} & \agg{8.0\%}{1.4\%} & \agg{9.2\%}{3.8\%} & \agg{5.5\%}{0.6\%} \\
  \quad succ (pass@1)   & \agg{10.0\%}{1.4\%} & \agg{9.5\%}{2.6\%} & \agg{9.0\%}{1.8\%} & \agg{6.2\%}{1.5\%} \\
  \midrule
  \cellcolor{cyan!20}\textbf{CoT+ToT+RAP data} {\scriptsize (quality: \textbf{87.6\%})} & & & & \\
  \quad succ (greedy)   & \agg{20.0\%}{2.7\%} & \agg{20.5\%}{3.7\%} & \agg{20.8\%}{2.1\%} & \agg{10.3\%}{3.6\%} \\
  \quad succ (pass@1)   & \agg{17.0\%}{2.2\%} & \agg{18.8\%}{1.3\%} & \agg{20.0\%}{3.4\%} & \agg{12.8\%}{3.6\%} \\
  \bottomrule
\end{tabular}
\vspace{0.5em}
\end{subtable}
\label{tab:cot_reasoning-unlearning}
\vspace{-1em}
\end{table}

\begin{table}[H]
    \footnotesize
    \centering
    \caption{\small\textbf{Additional results with alternative unlearning methods: CoT and search-based reasoning performance of LLMs fine-tuned and evaluated on Game-of-24.} Each cell shows the averaged result (4 seeds) of an LLM fine-tuned on data from specified sources. We highlight the best results for each pair of training data and inference method.\looseness=-1}
    \vspace{-1em}
\begin{subtable}{\linewidth}
\centering
\caption{Gradient Ascent}
\vspace{-0.5em}
\begin{tabular}{l | c c c c}
  \toprule
  & Learning rate & CoT (greedy) & ToT & RAP \\
  \midrule
  \textbf{Base LLM (Qwen2.5-Math 7B)} & N/A & 6.0\% & 28.0\% & 27.0\% \\
  \midrule
  \textbf{CoT data} {\scriptsize (quality: \textbf{92.9\%})} & 5e-6 & \agg{\textbf{26.8\%}}{2.1\%} & \agg{15.5\%}{4.1\%} & \agg{11.2\%}{1.7\%} \\
   & 2e-6 & \agg{23.0\%}{0.8\%} & \agg{\textbf{31.0\%}}{8.8\%} & \agg{22.8\%}{4.5\%} \\
   & 1e-6 & \agg{20.2\%}{1.0\%} & \agg{24.2\%}{3.5\%} & \agg{\textbf{28.0\%}}{2.6\%} \\
  \midrule
  \textbf{ToT+RAP data} {\scriptsize (quality: \textbf{69.3\%})} & 5e-6 & \agg{\textbf{13.5\%}}{5.1\%} & \agg{6.8\%}{2.9\%} & \agg{6.2\%}{2.2\%} \\
   & 2e-6 & \agg{9.8\%}{1.5\%} & \agg{20.0\%}{0.8\%} & \agg{\textbf{18.8\%}}{7.6\%} \\
   & 1e-6 & \agg{11.8\%}{1.7\%} & \agg{\textbf{20.8\%}}{5.9\%} & \agg{18.5\%}{3.3\%} \\
  \midrule
  \textbf{CoT+ToT+RAP data} {\scriptsize (quality: \textbf{95.3\%})} & 5e-6 & \agg{\textbf{28.5\%}}{1.7\%} & \agg{14.8\%}{2.5\%} & \agg{12.0\%}{1.8\%} \\
   & 2e-6 & \agg{20.2\%}{1.9\%} & \agg{\textbf{29.5\%}}{3.3\%} & \agg{\textbf{21.8\%}}{2.5\%} \\
   & 1e-6 & \agg{17.5\%}{1.3\%} & \agg{23.8\%}{3.8\%} & \agg{13.5\%}{3.1\%} \\
  \bottomrule
\end{tabular}
\vspace{0.5em}
\end{subtable}

\begin{subtable}{\linewidth}
\centering
\caption{RepNoise}
\vspace{-0.5em}
\begin{tabular}{l | c c c c}
  \toprule
  & Learning rate & CoT (greedy) & ToT & RAP \\
  \midrule
  \textbf{Base LLM (Qwen2.5-Math 7B)} & N/A & 6.0\% & 28.0\% & 27.0\% \\
  \midrule
  \textbf{CoT data} {\scriptsize (quality: \textbf{92.9\%})} & 5e-6 & \agg{\textbf{20.5\%}}{3.7\%} & \agg{0.0\%}{0.0\%} & \agg{0.0\%}{0.0\%} \\
   & 2e-6 & \agg{13.5\%}{0.6\%} & \agg{0.0\%}{0.0\%} & \agg{0.0\%}{0.0\%} \\
   & 1e-6 & \agg{8.0\%}{0.8\%} & \agg{0.0\%}{0.0\%} & \agg{0.0\%}{0.0\%} \\
  \midrule
  \textbf{ToT+RAP data} {\scriptsize (quality: \textbf{69.3\%})} & 5e-6 & \agg{\textbf{12.8\%}}{3.0\%} & \agg{0.0\%}{0.0\%} & \agg{0.0\%}{0.0\%} \\
   & 2e-6 & \agg{8.2\%}{1.3\%} & \agg{0.0\%}{0.0\%} & \agg{0.0\%}{0.0\%} \\
   & 1e-6 & \agg{6.0\%}{1.4\%} & \agg{0.0\%}{0.0\%} & \agg{0.0\%}{0.0\%} \\
  \midrule
  \textbf{CoT+ToT+RAP data} {\scriptsize (quality: \textbf{95.3\%})} & 5e-6 & \agg{\textbf{20.8\%}}{2.1\%} & \agg{0.0\%}{0.0\%} & \agg{0.0\%}{0.0\%} \\
   & 2e-6 & \agg{17.8\%}{1.0\%} & \agg{0.0\%}{0.0\%} & \agg{0.0\%}{0.0\%} \\
   & 1e-6 & \agg{14.5\%}{3.4\%} & \agg{0.0\%}{0.0\%} & \agg{0.0\%}{0.0\%} \\
  \bottomrule
\end{tabular}
\vspace{0.5em}
\end{subtable}
\label{tab:unlearning_methods_lr}
\vspace{-1em}
\end{table}

\section{Reasoner Details}
\label{sec:llm_reasoners}

\subsection{The Problems}
This section provides details on the two math games not covered in the main paper.

\paragraph{Game-of-24.}
We randomly split the first 900 cases (rank \#1-900 by human average performance in the dataset\footnote{\url{https://github.com/maitrix-org/llm-reasoners/blob/main/examples/ToT/game24/data/24.csv}}) into $\mathcal X_{\text{train}}$ (720 cases) and  $\mathcal X_{\text{valid}}$ (180 cases). The $\mathcal X_{\text{test}}$ are the next 100 cases (rank \#901-1000) following the setup in ToT~\citep{yao2023tree}. As a result, the validation set evaluates in-distribution (ID) generalization, while the test set evaluates out-of-distribution (OOD) generalization based on human difficulty levels. Each case is guaranteed to be solvable using the four input numbers, which range from 1 to 13.

\paragraph{Countdown.} We follow SoS codebase\footnote{\url{https://github.com/kanishkg/stream-of-search}}~\citep{gandhi2024stream} to randomly generate 500k training cases, 1k validation cases, and 1k test cases. The training and validation sets share the same distribution of target numbers, while the test set includes distinct target numbers (including the number 24). Each problem consists of four input numbers sampled from 1 to 99, with a target number between 10 and 100. All generated problems are guaranteed to have at least one valid solution path using only integer intermediate results.

\subsection{CoT Reasoner}

\begin{lstlisting}[caption={\small\textbf{Few-shot CoT template} on Countdown (5-shot prompt; only 2 shown for brevity), used in \textbf{data generation}. Here, \texttt{<input>} denotes a placeholder for the input numbers, \texttt{<target>} for the target number, and \texttt{<response>} for the LLM's output.},label={lst:cot-few-shot}]
Use numbers and basic arithmetic operations (+ - * /) to obtain the target number. Each step, you are only allowed to choose two of the remaining numbers to obtain a new number.

Input: 25 5 5 33 Target: 27
Steps:
25 + 5 = 30 (left: 5 33 30)
30 / 5 = 6 (left: 33 6)
33 - 6 = 27 (left: 27)
Answer: 33 - ((25 + 5) / 5) = 27

Input: 45 10 11 70 Target: 94
Steps:
10 + 11 = 21 (left: 45 70 21)
45 + 70 = 115 (left: 21 115)
115 - 21 = 94 (left: 94)
Answer: (45 + 70) - (10 + 11) = 94

Input: <input> Target: <target>
Steps: <response>
\end{lstlisting}

\begin{lstlisting}[caption={\small\textbf{Zero-shot CoT template} on Countdown, used in \textbf{fine-tuning and evaluation}.},label={lst:cot-zero-shot}]
Use numbers and basic arithmetic operations (+ - * /) to obtain the target number.

Input: <input> Target: <target>
Steps: <response>
\end{lstlisting}

Chain-of-Thought~\citep{wei2022chain} uses the few-shot prompting template (see \autoref{lst:cot-few-shot} for Countdown; Game-of-24 template follows the same format) for data generation and the zero-shot template (see \autoref{lst:cot-zero-shot}) for fine-tuning and evaluation. The ``greedy-decoding'' evaluation uses zero temperature. The ``pass@1'' evaluation uses a temperature of 0.7 and top\_p of 0.8 to sample 8 paths (n=8) for Countdown and 20 paths (n=20) for Game-of-24, then calculates the average success rate. 

\subsection{ToT and RAP Reasoners}

\begin{lstlisting}[caption={\small\textbf{Proposed prompt template} used for search-based reasoners on Countdown (6-shot prompt; only 3 shown for brevity).},label={lst:search-propose}]
Perform a basic arithmetic operation (+, -, *, /) on any two of the given numbers, replacing them with the result. Your goal is to explore combinations that may lead to a final result of the target number.

Input: 25 5 5 33 Target: 27
Possible next steps:
25 + 5 = 30 (left: 5 33 30)
25 * 5 = 125 (left: 5 33 125)
25 / 5 = 5 (left: 5 33 5)
25 + 33 = 58 (left: 5 5 58)
5 * 5 = 25 (left: 25 33 25)
5 + 33 = 38 (left: 25 5 38)
33 - 25 = 8 (left: 5 5 8)
33 - 5 = 28 (left: 25 5 28)

Input: 9 25 2 Target: 43
Possible next steps:
9 + 25 = 34 (left: 34 2)
9 * 2 = 18 (left: 25 18)
9 + 2 = 11 (left: 25 11)
25 * 2 = 50 (left: 9 50)
25 + 2 = 27 (left: 9 27)

Input: 21 115 Target: 94
Possible next steps:
115 - 21 = 94 (left: 94)
115 + 21 = 136 (left: 136)

Input: <input> Target: <target>
Possible next steps: <response>
\end{lstlisting}

\begin{lstlisting}[caption={\small\textbf{Intermediate reward prompt template} used for search-based reasoners on Countdown (11-shot prompt; only 6 shown for brevity).},label={lst:search-interreward}]
Evaluate if given number(s) can reach the target number (sure/likely/impossible)

Input: 27 Target: 27
sure

Input: 25 18 Target: 43
25 + 18 = 43
sure

Input: 31 10 Target: 52
31 + 10 = 41
31 - 10 = 21
31 * 10 = 310
31 / 10 = 3.1
impossible

Input: 45 70 21 Target: 94
45 + 70 + 21 = 115 + 21 = 136
-45 + 70 + 21 = 25 + 21 = 46
45 + 70 - 21 = 115 - 21 = 94
sure

Input: 15 16 16 Target: 43
15 + 16 + 16 = 47
(16 - 15) * 16 = 1 * 16 = 16
I cannot obtain 43 now, but numbers are within a reasonable range
likely

Input: 90 108 97 Target: 27
90 + 108 + 97 = 295
90 - 108 + 97 = 79
90 108 97 are all too big
impossible

Input: <input> Target: <target>
<response>
\end{lstlisting}

\begin{lstlisting}[caption={\small\textbf{Terminal reward prompt template} used for search-based reasoners on Countdown (6-shot prompt; only 4 shown for brevity). Here, \texttt{<answer>} is a given equation.},label={lst:search-terminalreward}]
Use numbers and basic arithmetic operations (+ - * /) to obtain the target number. Given an input and an answer, give a judgement (sure/impossible) if the answer is correct, i.e. it uses each input exactly once and no other numbers, and reach the target number.

Input: 25 5 5 33 Target: 27
Answer: 33 - ((25 + 5) / 5) = 27
Judge: sure

Input: 92 91 23 54 Target: 78
Answer: 92 + 91 - (23 + 54) = 106
Judge: impossible

Input: 45 13 11 70 Target: 94
Answer: 13 + 11 + 70 = 94
Judge: impossible

Input: 25 5 5 33 Target: 27
Answer: (33 - 5) * (25 / 5) = 27
Judge: impossible

Input: <input> Target: <target>
Answer: <answer>
Judge: <response>
\end{lstlisting}

The search-based reasoners (ToT~\citep{yao2023tree} and RAP~\citep{hao2023reasoning}) are implemented using the LLM-reasoners codebase~\citep{hao2024llm}. Both use the same prompt templates but differ in their underlying search algorithm. 

\paragraph{Tree-of-Thought~\citep{yao2023tree}.} ToT employs a beam search strategy. It begins with an empty set of nodes (the beam). At each search step, the LLM is prompted using the \textit{propose prompt} (\autoref{lst:search-propose}) to generate plausible next steps for each node in the beam. Each proposed step is then evaluated with the \textit{intermediate reward prompt} (\autoref{lst:search-interreward}), which labels responses as ``sure'', ``likely'', or ``impossible''. Following the original ToT setup, these labels are mapped to heuristic scores of 1, 0.1, and 0.0001, respectively. To reduce variance, we sample three responses for each reward step, and compute the average heuristic reward.

The beam is updated by selecting the top candidate nodes based on these heuristic scores, maintaining the predefined beam size. Since both Countdown and Game-of-24 tasks involve four steps in chain-of-thought reasoning, this process is repeated for four iterations. At the final search step, the \textit{terminal reward prompt} (\autoref{lst:search-terminalreward}) is used to assign a terminal reward to each node in the final beam. The path with the highest terminal reward is selected as the final reasoning path for evaluation, while all paths in the final beam are retained for training data. 

To improve efficiency, we optimize the codebase using vLLM~\citep{kwon2023efficient}, which enables batched inference across $\approx$1000 cases at the same time. In total, the ToT process requires $4 \times 2 = 8$ passes (i.e., 8 calls to \texttt{llm.generate()} in vLLM) per batch. This batched approach significantly accelerates the search process, achieving \textbf{over 100$\times$ speedup} compared to the original for-loop execution in the LLM-Reasoners codebase.

In practice, we vary the beam size between 5 and 16 while keeping all other search parameters fixed during training data generation. Empirically, we find that a beam size of 5 for Countdown and 6 for Game-of-24 yields the best performance on Qwen2.5-Math 7B, while beam sizes of 8 and 10 perform best for the respective tasks on Qwen2.5-Math 1.5B. Accordingly, we adopt these optimal beam sizes for evaluating both the base LLMs and the improved models fine-tuned from them. 

\paragraph{Reasoning-via-Planning~\citep{hao2023reasoning}.}
RAP employs a Monte Carlo Tree Search (MCTS) strategy, maintaining a search tree initialized as empty. At each iteration, a node is selected for expansion using the Upper Confidence Bound applied to Trees (UCT) algorithm~\citep{kocsis2006bandit}, where the exploration parameter balances exploitation and exploration. The selected node is then expanded by simulating a full rollout: the \textit{propose prompt} generates plausible next steps, and \textit{reward prompts} are used to assign scores to each step along the way. Steps are sampled proportionally to their rewards throughout the rollout. After completing the rollout, the terminal reward is backpropagated through the tree to update the value estimates of the visited nodes.

We run this process for 100 iterations, resulting in up to $100\times 4\times 2 = 800$ passes. This is a loose upper bound, as many selected nodes are leaf nodes or near-terminal, leading to shorter rollouts in practice. While the original LLM-Reasoners implementation uses only 10 iterations, we increase it to 100 to improve performance, made feasible by our use of batched inference. At the end of the search, the path with the highest terminal reward is selected for evaluation, and all explored paths are extracted for training data.

In practice, we vary the exploration parameter over (1.0, 2.0, 4.0, 6.0, 8.0, 10.0) while keeping all other search parameters fixed during training data generation. Empirically, we find that an exploration parameter of 1.0 yields the best performance for both Countdown and Game-of-24 on Qwen2.5-Math 7B, whereas a value of 2.0 performs best for both tasks on Qwen2.5-Math 1.5B. Accordingly, we adopt these optimal exploration parameters for evaluating both the base LLMs and the improved models fine-tuned from them. 

\subsection{Loss of Search Capability after CoT Fine-Tuning: A Planning Perspective}
\label{sec:explanation}

An important insight emerges when comparing the prompting strategies used in CoT reasoning versus search-based methods like ToT and RAP: although they aim to solve the same task, they frame the role of the LLM in fundamentally different ways.

CoT treats the LLM as an \textit{open-loop controller}, generating complete reasoning paths in a single pass (\autoref{lst:cot-few-shot}) without any feedback or evaluation. In contrast, search-based methods integrates the LLM into a \textit{closed-loop system}, where the LLM serves two interconnected roles: a policy that proposes next steps (\autoref{lst:search-propose}) and a reward model that evaluates them (\autoref{lst:search-interreward}, \autoref{lst:search-terminalreward}). This interaction between proposal and intermediate feedback is central to structured planning and search.

This difference has meaningful consequences for fine-tuning. When an LLM is fine-tuned on CoT-style paths that strip away reward signals, it may weaken its ability to perform reward modeling, which is critical in search-based reasoning. 
This observation offers a planning-theoretic explanation for why models fine-tuned on CoT-style paths may struggle with inference-time search.

\subsection{Classic BFS and DFS Reasoners}

We also employ classic search algorithms, Breadth-First Search (BFS) and Depth-First Search (DFS), to generate training data for the Countdown task. We adopt the implementation from SoS~\citep{gandhi2024stream} without modification. These symbolic reasoners are significantly faster than LLM reasoners and achieve high success rates of 65.4\% (BFS) and 81.5\% (DFS) on the training set (due to the use of pruning heuristics, they do not guarantee a solution in all cases). 
However, unlike LLM reasoners, these methods rely on external tools such as calculators and goal checkers to guide the search process. 

\section{Limitation} 

This paper focuses on improving \textit{reasoning language models}.  Preliminary experiments on one generic language model, Mistral 7B~\citep{jiang2023mistral}, show that while diverse reasoners still provide benefits, small learning rate and the forgetting objective offer limited improvement. We suspect these effects depend more on the base model's reasoning ability than size, similar to observations in RL that performance is bounded by base model capability~\citep{yue2025does}. Extending our method to other reasoning-oriented LLMs is a valuable direction for future work.

\section{Results on Standardized Math Benchmarks}
\label{app:math}

For completeness, we also report results on standardized math benchmarks. These evaluations are not the main focus of our work, which centers on math puzzles such as Countdown and Game-of-24.

The training problems for our standardized math experiments are drawn from the OpenThoughts-114k-math dataset\footnote{\url{https://huggingface.co/datasets/open-r1/OpenThoughts-114k-math}}\citep{guha2025openthoughts}, which itself is extracted from NuminaMath-CoT~\citep{li2024numinamath}. This corpus integrates problems from many high-school math competitions, including the Olympiad, MATH~\citep{hendrycksmath2021}, AoPS forums, and AMC/AIME competitions. From this collection, we curated a subset of 38k problems with \textit{numerical} gold answers, which we used as our training problems.

The training CoTs are constructed from three sources: (1) human-written CoTs containing gold answers in OpenThoughts, (2) self-generated CoTs whose final answers match the gold answers, and (3) self-generated CoTs whose final answers do not match the gold answers. Since OpenThoughts, like many other public datasets, does not include negative CoTs, we generate them ourselves. 
To verify correctness, we employ \texttt{math-verify}\footnote{\url{https://github.com/huggingface/Math-Verify}} to check the answers wrapped by \verb|\\boxed{}| in CoTs, following common practice. The instruction used for both training and testing is: \verb|Please reason step by step, and put your final answer within \\boxed{}|. 
SFT is trained only on CoTs (1) and (2), while the other methods are trained with CoTs (1)–(3). It is worth noting that although CoTs in category (2) contain gold answers, their reasoning steps may still be incorrect, since no process-based verifier exists for general math problems.

Due to computational resource limits, we train each method for 2 epochs with a context length of 2048, using 4 random seeds. The training time for the 7B models is around 15 hours on 8 A100 GPUs (40GB each).

We evaluate each fine-tuned LLM on the held-out test datasets using real math competitions AMC-12 (2023 and 2024)\footnote{\url{https://artofproblemsolving.com/wiki/index.php/AMC_12_Problems_and_Solutions}}. We do not include more advanced contests in our evaluation, as they typically require substantially longer context lengths (exceeding 4k tokens~\citep{yang2025qwen3}) and large-scale online RL~\citep{guo2025deepseek}, which go beyond the scope of offline fine-tuning considered in this work. 

\begin{table}[H]
    \centering
    \caption{\textbf{CoT reasoning performance of LLMs fine-tuned on 38k high-school math problems and evaluated on AMC-12 datasets.} Each method is trained with 4 seeds. The evaluation metrics are pass@1 success rate,  estimated from 32 sampled paths for each problem.\looseness=-1}
    \vspace{-1em}
\begin{subtable}{\linewidth}
\centering
\caption{Base LLM: Qwen2.5-Math 7B}
\vspace{-0.5em}
\begin{tabular}{l | c c c c c}
  \toprule
   & Base & SFT & SimPO & CPO-SimPO & \cellcolor{cyan!20}UFT \\
  \midrule
  AMC 2023  & 56.3\% & \agg{\textbf{61.5\%}}{0.9\%} & \agg{0\%}{0\%} & \agg{\textbf{59.3\%}}{2.2\%} & \agg{\textbf{61.6\%}}{0.8\%} \\
  AMC 2024   & 29.5\% & \agg{\textbf{30.1\%}}{0.7\%} & \agg{0\%}{0\%} & \agg{\textbf{30.6\%}}{2.2\%} & \agg{\textbf{29.8\%}}{0.6\%} \\
  \bottomrule
\end{tabular}
\vspace{0.5em}
\end{subtable}

\begin{subtable}{\linewidth}
\centering
\caption{Base LLM: Qwen2.5-Math 1.5B}
\vspace{-0.5em}
\begin{tabular}{l | c c c c c}
  \toprule
   & Base & SFT & SimPO & CPO-SimPO & \cellcolor{cyan!20}UFT \\
  \midrule
  AMC 2023  & 38.3\% & \agg{\textbf{50.0\%}}{2.2\%} & \agg{0\%}{0\%} & \agg{\textbf{50.9\%}}{3.2\%} & \agg{\textbf{50.2\%}}{1.2\%} \\
  AMC 2024   & 18.6\% & \agg{\textbf{28.8\%}}{0.5\%} & \agg{0\%}{0\%} & \agg{\textbf{28.6\%}}{0.6\%} & \agg{\textbf{28.7\%}}{0.5\%} \\
  \bottomrule
\end{tabular}
\vspace{0.5em}
\end{subtable}

    \label{tab:math}
\vspace{-1em}
\end{table}

\autoref{tab:math} shows the results. Except for SimPO, all other methods are not statistically distinguishable. Moreover, the improvements over the base LLM are minor for all 7B models. This likely reflects the noisier nature of human-written general math training data compared to the synthetic, structured data in our math puzzles. In math competitions, test problems differ substantially in style and content from training problems across years, making it harder to generalize from failed reasoning paths. By contrast, in our math puzzles, training and test problems share the same synthetic structure, which facilitates generalization. The weaker performance may also stem from our limited training setup (38k problems, 2048 context length). 
Nevertheless, unlike SimPO, our method UFT, which also leverages negative paths, remains as stable as SFT, underscoring its robustness under limited training conditions.

%% file: main.bbl
\begin{thebibliography}{95}
\providecommand{\natexlab}[1]{#1}
\providecommand{\url}[1]{\texttt{#1}}
\expandafter\ifx\csname urlstyle\endcsname\relax
  \providecommand{\doi}[1]{doi: #1}\else
  \providecommand{\doi}{doi: \begingroup \urlstyle{rm}\Url}\fi

\bibitem[Achiam et~al.(2023)Achiam, Adler, Agarwal, Ahmad, Akkaya, Aleman, Almeida, Altenschmidt, Altman, Anadkat, et~al.]{achiam2023gpt}
Josh Achiam, Steven Adler, Sandhini Agarwal, Lama Ahmad, Ilge Akkaya, Florencia~Leoni Aleman, Diogo Almeida, Janko Altenschmidt, Sam Altman, Shyamal Anadkat, et~al.
\newblock Gpt-4 technical report.
\newblock \emph{arXiv preprint arXiv:2303.08774}, 2023.

\bibitem[Bai et~al.(2022)Bai, Kadavath, Kundu, Askell, Kernion, Jones, Chen, Goldie, Mirhoseini, McKinnon, et~al.]{bai2022constitutional}
Yuntao Bai, Saurav Kadavath, Sandipan Kundu, Amanda Askell, Jackson Kernion, Andy Jones, Anna Chen, Anna Goldie, Azalia Mirhoseini, Cameron McKinnon, et~al.
\newblock Constitutional ai: Harmlessness from ai feedback.
\newblock \emph{arXiv preprint arXiv:2212.08073}, 2022.

\bibitem[Bohnet et~al.(2024)Bohnet, Nova, Parisi, Swersky, Goshvadi, Dai, Schuurmans, Fiedel, and Sedghi]{nova2024palnningBenchamrk}
Bernd Bohnet, Azade Nova, Aaron~T Parisi, Kevin Swersky, Katayoon Goshvadi, Hanjun Dai, Dale Schuurmans, Noah Fiedel, and Hanie Sedghi.
\newblock Exploring and benchmarking the planning capabilities of large language models, 2024.

\bibitem[Cen et~al.(2025)Cen, Liu, Zeng, Chaudhari, Rangwala, Karypis, and Fakoor]{cen2025bridging}
Zhepeng Cen, Yao Liu, Siliang Zeng, Pratik Chaudhari, Huzefa Rangwala, George Karypis, and Rasool Fakoor.
\newblock Bridging the training-inference gap in {LLM}s by leveraging self-generated tokens.
\newblock \emph{Transactions on Machine Learning Research}, 2025.
\newblock ISSN 2835-8856.

\bibitem[Chen et~al.(2025)Chen, Dong, Wei, Huang, Zhang, Su, and Zhu]{chen2025understanding}
Huanran Chen, Yinpeng Dong, Zeming Wei, Yao Huang, Yichi Zhang, Hang Su, and Jun Zhu.
\newblock Understanding pre-training and fine-tuning from loss landscape perspectives.
\newblock \emph{arXiv preprint arXiv:2505.17646}, 2025.

\bibitem[Chen et~al.(2021)Chen, Tworek, Jun, Yuan, Pinto, Kaplan, Edwards, Burda, Joseph, Brockman, et~al.]{chen2021evaluating}
Mark Chen, Jerry Tworek, Heewoo Jun, Qiming Yuan, Henrique Ponde De~Oliveira Pinto, Jared Kaplan, Harri Edwards, Yuri Burda, Nicholas Joseph, Greg Brockman, et~al.
\newblock Evaluating large language models trained on code.
\newblock \emph{arXiv preprint arXiv:2107.03374}, 2021.

\bibitem[Chen et~al.(2024{\natexlab{a}})Chen, Wang, Wu, Chen, Xu, Luo, Zhang, and Zhang]{chen2024advancing}
Sijia Chen, Yibo Wang, Yi-Feng Wu, Qingguo Chen, Zhao Xu, Weihua Luo, Kaifu Zhang, and Lijun Zhang.
\newblock Advancing tool-augmented large language models: Integrating insights from errors in inference trees.
\newblock \emph{Advances in Neural Information Processing Systems}, 37:\penalty0 106555--106581, 2024{\natexlab{a}}.

\bibitem[Chen et~al.(2024{\natexlab{b}})Chen, Xu, Liang, He, Pang, Yu, Song, Liu, Zhou, Zhang, et~al.]{chen2024not}
Xingyu Chen, Jiahao Xu, Tian Liang, Zhiwei He, Jianhui Pang, Dian Yu, Linfeng Song, Qiuzhi Liu, Mengfei Zhou, Zhuosheng Zhang, et~al.
\newblock Do not think that much for 2+ 3=? on the overthinking of o1-like llms.
\newblock \emph{arXiv preprint arXiv:2412.21187}, 2024{\natexlab{b}}.

\bibitem[Chen et~al.(2024{\natexlab{c}})Chen, White, Mooney, Payani, Su, and Sun]{chen2024tree}
Ziru Chen, Michael White, Raymond Mooney, Ali Payani, Yu~Su, and Huan Sun.
\newblock When is tree search useful for llm planning? it depends on the discriminator.
\newblock \emph{arXiv preprint arXiv:2402.10890}, 2024{\natexlab{c}}.

\bibitem[Chollet et~al.(2024)Chollet, Knoop, Kamradt, and Landers]{chollet2024arc}
Francois Chollet, Mike Knoop, Gregory Kamradt, and Bryan Landers.
\newblock Arc prize 2024: Technical report.
\newblock \emph{arXiv preprint arXiv:2412.04604}, 2024.

\bibitem[Cobbe et~al.(2021)Cobbe, Kosaraju, Bavarian, Chen, Jun, Kaiser, Plappert, Tworek, Hilton, Nakano, et~al.]{cobbe2021training}
Karl Cobbe, Vineet Kosaraju, Mohammad Bavarian, Mark Chen, Heewoo Jun, Lukasz Kaiser, Matthias Plappert, Jerry Tworek, Jacob Hilton, Reiichiro Nakano, et~al.
\newblock Training verifiers to solve math word problems.
\newblock \emph{arXiv preprint arXiv:2110.14168}, 2021.

\bibitem[Du et~al.(2023)Du, Li, Torralba, Tenenbaum, and Mordatch]{du2023improving}
Yilun Du, Shuang Li, Antonio Torralba, Joshua~B Tenenbaum, and Igor Mordatch.
\newblock Improving factuality and reasoning in language models through multiagent debate.
\newblock \emph{arXiv preprint arXiv:2305.14325}, 2023.

\bibitem[Ethayarajh et~al.(2024)Ethayarajh, Xu, Muennighoff, Jurafsky, and Kiela]{ethayarajh2024kto}
Kawin Ethayarajh, Winnie Xu, Niklas Muennighoff, Dan Jurafsky, and Douwe Kiela.
\newblock Kto: Model alignment as prospect theoretic optimization.
\newblock \emph{arXiv preprint arXiv:2402.01306}, 2024.

\bibitem[Fakoor et~al.(2020)Fakoor, Chaudhari, and Smola]{fakoor20a}
Rasool Fakoor, Pratik Chaudhari, and Alexander~J. Smola.
\newblock P3o: Policy-on policy-off policy optimization.
\newblock In \emph{Proceedings of The 35th Uncertainty in Artificial Intelligence Conference}, volume 115 of \emph{Proceedings of Machine Learning Research}, pp.\  1017--1027. PMLR, 22--25 Jul 2020.

\bibitem[Feng et~al.(2023)Feng, Wan, Wen, McAleer, Wen, Zhang, and Wang]{feng2023alphazero}
Xidong Feng, Ziyu Wan, Muning Wen, Stephen~Marcus McAleer, Ying Wen, Weinan Zhang, and Jun Wang.
\newblock Alphazero-like tree-search can guide large language model decoding and training.
\newblock \emph{arXiv preprint arXiv:2309.17179}, 2023.

\bibitem[Gandhi et~al.(2024)Gandhi, Lee, Grand, Liu, Cheng, Sharma, and Goodman]{gandhi2024stream}
Kanishk Gandhi, Denise Lee, Gabriel Grand, Muxin Liu, Winson Cheng, Archit Sharma, and Noah~D Goodman.
\newblock Stream of search (sos): Learning to search in language.
\newblock \emph{arXiv preprint arXiv:2404.03683}, 2024.

\bibitem[Gao et~al.(2024)Gao, Niu, He, Xu, Liu, Liu, Hu, and Wen]{gao2024interpretable}
Zitian Gao, Boye Niu, Xuzheng He, Haotian Xu, Hongzhang Liu, Aiwei Liu, Xuming Hu, and Lijie Wen.
\newblock Interpretable contrastive monte carlo tree search reasoning.
\newblock \emph{arXiv preprint arXiv:2410.01707}, 2024.

\bibitem[Glazer et~al.(2024)Glazer, Erdil, Besiroglu, Chicharro, Chen, Gunning, Olsson, Denain, Ho, Santos, et~al.]{glazer2024frontiermath}
Elliot Glazer, Ege Erdil, Tamay Besiroglu, Diego Chicharro, Evan Chen, Alex Gunning, Caroline~Falkman Olsson, Jean-Stanislas Denain, Anson Ho, Emily de~Oliveira Santos, et~al.
\newblock Frontiermath: A benchmark for evaluating advanced mathematical reasoning in ai.
\newblock \emph{arXiv preprint arXiv:2411.04872}, 2024.

\bibitem[Guha et~al.(2025)Guha, Marten, Keh, Raoof, Smyrnis, Bansal, Nezhurina, Mercat, Vu, Sprague, et~al.]{guha2025openthoughts}
Etash Guha, Ryan Marten, Sedrick Keh, Negin Raoof, Georgios Smyrnis, Hritik Bansal, Marianna Nezhurina, Jean Mercat, Trung Vu, Zayne Sprague, et~al.
\newblock Openthoughts: Data recipes for reasoning models.
\newblock \emph{arXiv preprint arXiv:2506.04178}, 2025.

\bibitem[Gulcehre et~al.(2023)Gulcehre, Paine, Srinivasan, Konyushkova, Weerts, Sharma, Siddhant, Ahern, Wang, Gu, et~al.]{gulcehre2023reinforced}
Caglar Gulcehre, Tom~Le Paine, Srivatsan Srinivasan, Ksenia Konyushkova, Lotte Weerts, Abhishek Sharma, Aditya Siddhant, Alex Ahern, Miaosen Wang, Chenjie Gu, et~al.
\newblock Reinforced self-training (rest) for language modeling.
\newblock \emph{arXiv preprint arXiv:2308.08998}, 2023.

\bibitem[Guo et~al.(2025)Guo, Yang, Zhang, Song, Zhang, Xu, Zhu, Ma, Wang, Bi, et~al.]{guo2025deepseek}
Daya Guo, Dejian Yang, Haowei Zhang, Junxiao Song, Ruoyu Zhang, Runxin Xu, Qihao Zhu, Shirong Ma, Peiyi Wang, Xiao Bi, et~al.
\newblock Deepseek-r1: Incentivizing reasoning capability in llms via reinforcement learning.
\newblock \emph{arXiv preprint arXiv:2501.12948}, 2025.

\bibitem[Hao et~al.(2023)Hao, Gu, Ma, Hong, Wang, Wang, and Hu]{hao2023reasoning}
Shibo Hao, Yi~Gu, Haodi Ma, Joshua~Jiahua Hong, Zhen Wang, Daisy~Zhe Wang, and Zhiting Hu.
\newblock Reasoning with language model is planning with world model.
\newblock \emph{arXiv preprint arXiv:2305.14992}, 2023.

\bibitem[Hao et~al.(2024)Hao, Gu, Luo, Liu, Shao, Wang, Xie, Ma, Samavedhi, Gao, et~al.]{hao2024llm}
Shibo Hao, Yi~Gu, Haotian Luo, Tianyang Liu, Xiyan Shao, Xinyuan Wang, Shuhua Xie, Haodi Ma, Adithya Samavedhi, Qiyue Gao, et~al.
\newblock Llm reasoners: New evaluation, library, and analysis of step-by-step reasoning with large language models.
\newblock \emph{arXiv preprint arXiv:2404.05221}, 2024.

\bibitem[Hendrycks et~al.(2021)Hendrycks, Burns, Kadavath, Arora, Basart, Tang, Song, and Steinhardt]{hendrycksmath2021}
Dan Hendrycks, Collin Burns, Saurav Kadavath, Akul Arora, Steven Basart, Eric Tang, Dawn Song, and Jacob Steinhardt.
\newblock Measuring mathematical problem solving with the math dataset.
\newblock \emph{NeurIPS}, 2021.

\bibitem[Hinton et~al.(2015)Hinton, Vinyals, and Dean]{hinton2015distilling}
Geoffrey Hinton, Oriol Vinyals, and Jeff Dean.
\newblock Distilling the knowledge in a neural network.
\newblock \emph{arXiv preprint arXiv:1503.02531}, 2015.

\bibitem[Hong et~al.(2024)Hong, Lee, and Thorne]{hong2024orpo}
Jiwoo Hong, Noah Lee, and James Thorne.
\newblock Orpo: Monolithic preference optimization without reference model.
\newblock \emph{arXiv preprint arXiv:2403.07691}, 2024.

\bibitem[Hosseini et~al.(2024)Hosseini, Yuan, Malkin, Courville, Sordoni, and Agarwal]{hosseini2024v}
Arian Hosseini, Xingdi Yuan, Nikolay Malkin, Aaron Courville, Alessandro Sordoni, and Rishabh Agarwal.
\newblock V-star: Training verifiers for self-taught reasoners.
\newblock \emph{arXiv preprint arXiv:2402.06457}, 2024.

\bibitem[Huang et~al.(2024)Huang, Hu, Ilhan, Tekin, and Liu]{huang2024booster}
Tiansheng Huang, Sihao Hu, Fatih Ilhan, Selim~Furkan Tekin, and Ling Liu.
\newblock Booster: Tackling harmful fine-tuning for large language models via attenuating harmful perturbation.
\newblock \emph{arXiv preprint arXiv:2409.01586}, 2024.

\bibitem[Jiang et~al.(2023)Jiang, Sablayrolles, Mensch, Bamford, Chaplot, de~las Casas, Bressand, Lengyel, Lample, Saulnier, et~al.]{jiang2023mistral}
AQ~Jiang, A~Sablayrolles, A~Mensch, C~Bamford, DS~Chaplot, D~de~las Casas, F~Bressand, G~Lengyel, G~Lample, L~Saulnier, et~al.
\newblock Mistral 7b (2023).
\newblock \emph{arXiv preprint arXiv:2310.06825}, 2023.

\bibitem[Keskar et~al.(2019)Keskar, McCann, Varshney, Xiong, and Socher]{keskar2019ctrl}
Nitish~Shirish Keskar, Bryan McCann, Lav~R Varshney, Caiming Xiong, and Richard Socher.
\newblock Ctrl: A conditional transformer language model for controllable generation.
\newblock \emph{arXiv preprint arXiv:1909.05858}, 2019.

\bibitem[Kocsis \& Szepesv{\'a}ri(2006)Kocsis and Szepesv{\'a}ri]{kocsis2006bandit}
Levente Kocsis and Csaba Szepesv{\'a}ri.
\newblock Bandit based monte-carlo planning.
\newblock In \emph{European conference on machine learning}, pp.\  282--293. Springer, 2006.

\bibitem[Koh et~al.(2024)Koh, McAleer, Fried, and Salakhutdinov]{koh2024tree}
Jing~Yu Koh, Stephen McAleer, Daniel Fried, and Ruslan Salakhutdinov.
\newblock Tree search for language model agents.
\newblock \emph{arXiv preprint arXiv:2407.01476}, 2024.

\bibitem[Kwon et~al.(2023)Kwon, Li, Zhuang, Sheng, Zheng, Yu, Gonzalez, Zhang, and Stoica]{kwon2023efficient}
Woosuk Kwon, Zhuohan Li, Siyuan Zhuang, Ying Sheng, Lianmin Zheng, Cody~Hao Yu, Joseph Gonzalez, Hao Zhang, and Ion Stoica.
\newblock Efficient memory management for large language model serving with pagedattention.
\newblock In \emph{Proceedings of the 29th Symposium on Operating Systems Principles}, pp.\  611--626, 2023.

\bibitem[Lehnert et~al.(2024)Lehnert, Sukhbaatar, Su, Zheng, Mcvay, Rabbat, and Tian]{lehnert2024beyond}
Lucas Lehnert, Sainbayar Sukhbaatar, DiJia Su, Qinqing Zheng, Paul Mcvay, Michael Rabbat, and Yuandong Tian.
\newblock Beyond a*: Better planning with transformers via search dynamics bootstrapping.
\newblock \emph{arXiv preprint arXiv:2402.14083}, 2024.

\bibitem[Li et~al.(2024)Li, Beeching, Tunstall, Lipkin, Soletskyi, Huang, Rasul, Yu, Jiang, Shen, et~al.]{li2024numinamath}
Jia Li, Edward Beeching, Lewis Tunstall, Ben Lipkin, Roman Soletskyi, Shengyi Huang, Kashif Rasul, Longhui Yu, Albert~Q Jiang, Ziju Shen, et~al.
\newblock Numinamath: The largest public dataset in ai4maths with 860k pairs of competition math problems and solutions.
\newblock \emph{Hugging Face repository}, 13\penalty0 (9):\penalty0 9, 2024.

\bibitem[Lightman et~al.(2023)Lightman, Kosaraju, Burda, Edwards, Baker, Lee, Leike, Schulman, Sutskever, and Cobbe]{lightman2023let}
Hunter Lightman, Vineet Kosaraju, Yura Burda, Harri Edwards, Bowen Baker, Teddy Lee, Jan Leike, John Schulman, Ilya Sutskever, and Karl Cobbe.
\newblock Let's verify step by step.
\newblock \emph{arXiv preprint arXiv:2305.20050}, 2023.

\bibitem[Liu et~al.(2024)Liu, Wei, Liu, Si, Zhang, Rao, Zheng, Peng, Yang, Zhou, et~al.]{liu2024best}
Ruibo Liu, Jerry Wei, Fangyu Liu, Chenglei Si, Yanzhe Zhang, Jinmeng Rao, Steven Zheng, Daiyi Peng, Diyi Yang, Denny Zhou, et~al.
\newblock Best practices and lessons learned on synthetic data.
\newblock \emph{arXiv preprint arXiv:2404.07503}, 2024.

\bibitem[Lopez-Paz \& Ranzato(2017)Lopez-Paz and Ranzato]{lopez2017gradient}
David Lopez-Paz and Marc'Aurelio Ranzato.
\newblock Gradient episodic memory for continual learning.
\newblock \emph{Advances in neural information processing systems}, 30, 2017.

\bibitem[Madaan et~al.(2024)Madaan, Tandon, Gupta, Hallinan, Gao, Wiegreffe, Alon, Dziri, Prabhumoye, Yang, et~al.]{madaan2024self}
Aman Madaan, Niket Tandon, Prakhar Gupta, Skyler Hallinan, Luyu Gao, Sarah Wiegreffe, Uri Alon, Nouha Dziri, Shrimai Prabhumoye, Yiming Yang, et~al.
\newblock Self-refine: Iterative refinement with self-feedback.
\newblock \emph{Advances in Neural Information Processing Systems}, 36, 2024.

\bibitem[McCloskey \& Cohen(1989)McCloskey and Cohen]{mccloskey1989catastrophic}
Michael McCloskey and Neal~J Cohen.
\newblock Catastrophic interference in connectionist networks: The sequential learning problem.
\newblock In \emph{Psychology of learning and motivation}, volume~24, pp.\  109--165. Elsevier, 1989.

\bibitem[Meng et~al.(2024{\natexlab{a}})Meng, Wang, Yang, Peng, and Chang]{meng2024llm}
Silin Meng, Yiwei Wang, Cheng-Fu Yang, Nanyun Peng, and Kai-Wei Chang.
\newblock Llm-a*: Large language model enhanced incremental heuristic search on path planning.
\newblock \emph{arXiv preprint arXiv:2407.02511}, 2024{\natexlab{a}}.

\bibitem[Meng et~al.(2024{\natexlab{b}})Meng, Xia, and Chen]{meng2024simpo}
Yu~Meng, Mengzhou Xia, and Danqi Chen.
\newblock Simpo: Simple preference optimization with a reference-free reward.
\newblock \emph{Advances in Neural Information Processing Systems}, 37:\penalty0 124198--124235, 2024{\natexlab{b}}.

\bibitem[Mirzadeh et~al.(2020)Mirzadeh, Farajtabar, Pascanu, and Ghasemzadeh]{mirzadeh2020understanding}
Seyed~Iman Mirzadeh, Mehrdad Farajtabar, Razvan Pascanu, and Hassan Ghasemzadeh.
\newblock Understanding the role of training regimes in continual learning.
\newblock \emph{Advances in Neural Information Processing Systems}, 33:\penalty0 7308--7320, 2020.

\bibitem[Mitchener et~al.(2025)Mitchener, Laurent, Tenmann, Narayanan, Wellawatte, White, Sani, and Rodriques]{mitchener2025bixbench}
Ludovico Mitchener, Jon~M Laurent, Benjamin Tenmann, Siddharth Narayanan, Geemi~P Wellawatte, Andrew White, Lorenzo Sani, and Samuel~G Rodriques.
\newblock Bixbench: a comprehensive benchmark for llm-based agents in computational biology.
\newblock \emph{arXiv preprint arXiv:2503.00096}, Feb 2025.
\newblock \doi{10.48550/arXiv.2503.00096}.
\newblock URL \url{https://arxiv.org/abs/2503.00096}.

\bibitem[Nakano et~al.(2021)Nakano, Hilton, Balaji, Wu, Ouyang, Kim, Hesse, Jain, Kosaraju, Saunders, et~al.]{nakano2021webgpt}
Reiichiro Nakano, Jacob Hilton, Suchir Balaji, Jeff Wu, Long Ouyang, Christina Kim, Christopher Hesse, Shantanu Jain, Vineet Kosaraju, William Saunders, et~al.
\newblock Webgpt: Browser-assisted question-answering with human feedback.
\newblock \emph{arXiv preprint arXiv:2112.09332}, 2021.

\bibitem[Ouyang et~al.(2022)Ouyang, Wu, Jiang, Almeida, Wainwright, Mishkin, Zhang, Agarwal, Slama, Ray, Schulman, Hilton, Kelton, Miller, Simens, Askell, Welinder, Christiano, Leike, and Lowe]{ouyang2022training}
Long Ouyang, Jeff Wu, Xu~Jiang, Diogo Almeida, Carroll~L. Wainwright, Pamela Mishkin, Chong Zhang, Sandhini Agarwal, Katarina Slama, Alex Ray, John Schulman, Jacob Hilton, Fraser Kelton, Luke Miller, Maddie Simens, Amanda Askell, Peter Welinder, Paul Christiano, Jan Leike, and Ryan Lowe.
\newblock Training language models to follow instructions with human feedback, 2022.

\bibitem[Pal et~al.(2024)Pal, Karkhanis, Dooley, Roberts, Naidu, and White]{pal2024smaug}
Arka Pal, Deep Karkhanis, Samuel Dooley, Manley Roberts, Siddartha Naidu, and Colin White.
\newblock Smaug: Fixing failure modes of preference optimisation with dpo-positive.
\newblock \emph{arXiv preprint arXiv:2402.13228}, 2024.

\bibitem[Pang et~al.(2024)Pang, Yuan, He, Cho, Sukhbaatar, and Weston]{pang2024iterative}
Richard~Yuanzhe Pang, Weizhe Yuan, He~He, Kyunghyun Cho, Sainbayar Sukhbaatar, and Jason Weston.
\newblock Iterative reasoning preference optimization.
\newblock \emph{Advances in Neural Information Processing Systems}, 37:\penalty0 116617--116637, 2024.

\bibitem[Qin et~al.(2023)Qin, Liang, Ye, Zhu, Yan, Lu, Lin, Cong, Tang, Qian, et~al.]{qin2023toolllm}
Yujia Qin, Shihao Liang, Yining Ye, Kunlun Zhu, Lan Yan, Yaxi Lu, Yankai Lin, Xin Cong, Xiangru Tang, Bill Qian, et~al.
\newblock Toolllm: Facilitating large language models to master 16000+ real-world apis.
\newblock \emph{arXiv preprint arXiv:2307.16789}, 2023.

\bibitem[Rafailov et~al.(2023)Rafailov, Sharma, Mitchell, Manning, Ermon, and Finn]{rafailov2023direct}
Rafael Rafailov, Archit Sharma, Eric Mitchell, Christopher~D Manning, Stefano Ermon, and Chelsea Finn.
\newblock Direct preference optimization: Your language model is secretly a reward model.
\newblock \emph{Advances in Neural Information Processing Systems}, 36:\penalty0 53728--53741, 2023.

\bibitem[Rosati et~al.(2024)Rosati, Wehner, Williams, Bartoszcze, Gonzales, Majumdar, Sajjad, Rudzicz, et~al.]{rosati2024representation}
Domenic Rosati, Jan Wehner, Kai Williams, Lukasz Bartoszcze, Robie Gonzales, Subhabrata Majumdar, Hassan Sajjad, Frank Rudzicz, et~al.
\newblock Representation noising: A defence mechanism against harmful finetuning.
\newblock \emph{Advances in Neural Information Processing Systems}, 37:\penalty0 12636--12676, 2024.

\bibitem[Setlur et~al.(2024)Setlur, Garg, Geng, Garg, Smith, and Kumar]{setlur2024rl}
Amrith Setlur, Saurabh Garg, Xinyang Geng, Naman Garg, Virginia Smith, and Aviral Kumar.
\newblock Rl on incorrect synthetic data scales the efficiency of llm math reasoning by eight-fold.
\newblock \emph{Advances in Neural Information Processing Systems}, 37:\penalty0 43000--43031, 2024.

\bibitem[Shinn et~al.(2023)Shinn, Cassano, Gopinath, Narasimhan, and Yao]{shinn2023reflexion}
Noah Shinn, Federico Cassano, Ashwin Gopinath, Karthik Narasimhan, and Shunyu Yao.
\newblock Reflexion: Language agents with verbal reinforcement learning.
\newblock \emph{Advances in Neural Information Processing Systems}, 36:\penalty0 8634--8652, 2023.

\bibitem[Singh et~al.(2023)Singh, Co-Reyes, Agarwal, Anand, Patil, Garcia, Liu, Harrison, Lee, Xu, et~al.]{singh2023beyond}
Avi Singh, John~D Co-Reyes, Rishabh Agarwal, Ankesh Anand, Piyush Patil, Xavier Garcia, Peter~J Liu, James Harrison, Jaehoon Lee, Kelvin Xu, et~al.
\newblock Beyond human data: Scaling self-training for problem-solving with language models.
\newblock \emph{arXiv preprint arXiv:2312.06585}, 2023.

\bibitem[Snell et~al.(2024)Snell, Lee, Xu, and Kumar]{snell2024scaling}
Charlie Snell, Jaehoon Lee, Kelvin Xu, and Aviral Kumar.
\newblock Scaling llm test-time compute optimally can be more effective than scaling model parameters.
\newblock \emph{arXiv preprint arXiv:2408.03314}, 2024.

\bibitem[Stechly et~al.(2024)Stechly, Valmeekam, and Kambhampati]{stechly2024chain}
Kaya Stechly, Karthik Valmeekam, and Subbarao Kambhampati.
\newblock Chain of thoughtlessness: An analysis of cot in planning.
\newblock \emph{arXiv preprint arXiv:2405.04776}, 2024.

\bibitem[Sutton et~al.(1998)Sutton, Barto, et~al.]{sutton1998reinforcement}
Richard~S Sutton, Andrew~G Barto, et~al.
\newblock \emph{Reinforcement learning: An introduction}, volume~1.
\newblock MIT press Cambridge, 1998.

\bibitem[Tamirisa et~al.(2024)Tamirisa, Bharathi, Phan, Zhou, Gatti, Suresh, Lin, Wang, Wang, Arel, et~al.]{tamirisa2024tamper}
Rishub Tamirisa, Bhrugu Bharathi, Long Phan, Andy Zhou, Alice Gatti, Tarun Suresh, Maxwell Lin, Justin Wang, Rowan Wang, Ron Arel, et~al.
\newblock Tamper-resistant safeguards for open-weight llms.
\newblock \emph{arXiv preprint arXiv:2408.00761}, 2024.

\bibitem[Tian et~al.(2024)Tian, Peng, Song, Jin, Yu, Han, Mi, and Yu]{tian2024toward}
Ye~Tian, Baolin Peng, Linfeng Song, Lifeng Jin, Dian Yu, Lei Han, Haitao Mi, and Dong Yu.
\newblock Toward self-improvement of llms via imagination, searching, and criticizing.
\newblock \emph{Advances in Neural Information Processing Systems}, 37:\penalty0 52723--52748, 2024.

\bibitem[Tuan \& Wang(2024)Tuan and Wang]{tuan2024gradient}
Yi-Lin Tuan and William~Yang Wang.
\newblock A gradient analysis framework for rewarding good and penalizing bad examples in language models.
\newblock \emph{arXiv preprint arXiv:2408.16751}, 2024.

\bibitem[Tunstall et~al.(2024)Tunstall, Beeching, Lambert, Rajani, Huang, Rasul, Bartolome, M.~Rush, and Wolf]{Tunstall_The_Alignment_Handbook}
Lewis Tunstall, Edward Beeching, Nathan Lambert, Nazneen Rajani, Shengyi Huang, Kashif Rasul, Alvaro Bartolome, Alexander M.~Rush, and Thomas Wolf.
\newblock {The Alignment Handbook}, 2024.
\newblock URL \url{https://github.com/huggingface/alignment-handbook}.

\bibitem[Uesato et~al.(2022)Uesato, Kushman, Kumar, Song, Siegel, Wang, Creswell, Irving, and Higgins]{uesato2022solving}
Jonathan Uesato, Nate Kushman, Ramana Kumar, Francis Song, Noah Siegel, Lisa Wang, Antonia Creswell, Geoffrey Irving, and Irina Higgins.
\newblock Solving math word problems with process-and outcome-based feedback.
\newblock \emph{arXiv preprint arXiv:2211.14275}, 2022.

\bibitem[Valmeekam et~al.(2023)Valmeekam, Marquez, Sreedharan, and Kambhampati]{valmeekam2023planning}
Karthik Valmeekam, Matthew Marquez, Sarath Sreedharan, and Subbarao Kambhampati.
\newblock On the planning abilities of large language models-a critical investigation.
\newblock \emph{Advances in Neural Information Processing Systems}, 36:\penalty0 75993--76005, 2023.

\bibitem[Vaswani et~al.(2017)Vaswani, Shazeer, Parmar, Uszkoreit, Jones, Gomez, Kaiser, and Polosukhin]{vaswani2017attention}
Ashish Vaswani, Noam Shazeer, Niki Parmar, Jakob Uszkoreit, Llion Jones, Aidan~N Gomez, {\L}ukasz Kaiser, and Illia Polosukhin.
\newblock Attention is all you need.
\newblock \emph{Advances in neural information processing systems}, 30, 2017.

\bibitem[von Werra et~al.(2020)von Werra, Belkada, Tunstall, Beeching, Thrush, Lambert, Huang, Rasul, and Gallouédec]{vonwerra2022trl}
Leandro von Werra, Younes Belkada, Lewis Tunstall, Edward Beeching, Tristan Thrush, Nathan Lambert, Shengyi Huang, Kashif Rasul, and Quentin Gallouédec.
\newblock Trl: Transformer reinforcement learning.
\newblock \url{https://github.com/huggingface/trl}, 2020.

\bibitem[Wang et~al.(2025)Wang, Fan, Zhang, Jia, Wei, Ram, Baracaldo, and Liu]{wang2025reasoning}
Changsheng Wang, Chongyu Fan, Yihua Zhang, Jinghan Jia, Dennis Wei, Parikshit Ram, Nathalie Baracaldo, and Sijia Liu.
\newblock Reasoning model unlearning: Forgetting traces, not just answers, while preserving reasoning skills.
\newblock \emph{arXiv preprint arXiv:2506.12963}, 2025.

\bibitem[Wang et~al.(2022)Wang, Wei, Schuurmans, Le, Chi, Narang, Chowdhery, and Zhou]{wang2022self}
Xuezhi Wang, Jason Wei, Dale Schuurmans, Quoc Le, Ed~Chi, Sharan Narang, Aakanksha Chowdhery, and Denny Zhou.
\newblock Self-consistency improves chain of thought reasoning in language models.
\newblock \emph{arXiv preprint arXiv:2203.11171}, 2022.

\bibitem[Wei et~al.(2022)Wei, Wang, Schuurmans, Bosma, Xia, Chi, Le, Zhou, et~al.]{wei2022chain}
Jason Wei, Xuezhi Wang, Dale Schuurmans, Maarten Bosma, Fei Xia, Ed~Chi, Quoc~V Le, Denny Zhou, et~al.
\newblock Chain-of-thought prompting elicits reasoning in large language models.
\newblock \emph{Advances in neural information processing systems}, 35:\penalty0 24824--24837, 2022.

\bibitem[Welleck et~al.(2019)Welleck, Kulikov, Roller, Dinan, Cho, and Weston]{welleck2019neural}
Sean Welleck, Ilia Kulikov, Stephen Roller, Emily Dinan, Kyunghyun Cho, and Jason Weston.
\newblock Neural text generation with unlikelihood training.
\newblock \emph{arXiv preprint arXiv:1908.04319}, 2019.

\bibitem[Williams(1992)]{williams1992simple}
Ronald~J Williams.
\newblock Simple statistical gradient-following algorithms for connectionist reinforcement learning.
\newblock \emph{Machine learning}, 8:\penalty0 229--256, 1992.

\bibitem[Xie et~al.(2023)Xie, Kawaguchi, Zhao, Zhao, Kan, He, and Xie]{xie2023self}
Yuxi Xie, Kenji Kawaguchi, Yiran Zhao, James~Xu Zhao, Min-Yen Kan, Junxian He, and Michael Xie.
\newblock Self-evaluation guided beam search for reasoning.
\newblock \emph{Advances in Neural Information Processing Systems}, 36:\penalty0 41618--41650, 2023.

\bibitem[Xie et~al.(2024)Xie, Goyal, Zheng, Kan, Lillicrap, Kawaguchi, and Shieh]{xie2024monte}
Yuxi Xie, Anirudh Goyal, Wenyue Zheng, Min-Yen Kan, Timothy~P Lillicrap, Kenji Kawaguchi, and Michael Shieh.
\newblock Monte carlo tree search boosts reasoning via iterative preference learning.
\newblock \emph{arXiv preprint arXiv:2405.00451}, 2024.

\bibitem[Xu(2024)]{xu2024cpo-simpo}
Haoran Xu.
\newblock The joint of contrastive preference optimization (cpo) \& simple preference optimization (simpo), 2024.
\newblock URL \url{https://github.com/fe1ixxu/CPO_SIMPO}.

\bibitem[Xu et~al.(2024)Xu, Sharaf, Chen, Tan, Shen, Van~Durme, Murray, and Kim]{xu2024contrastive}
Haoran Xu, Amr Sharaf, Yunmo Chen, Weiting Tan, Lingfeng Shen, Benjamin Van~Durme, Kenton Murray, and Young~Jin Kim.
\newblock Contrastive preference optimization: Pushing the boundaries of llm performance in machine translation.
\newblock \emph{arXiv preprint arXiv:2401.08417}, 2024.

\bibitem[Yang et~al.(2024)Yang, Zhang, Hui, Gao, Yu, Li, Liu, Tu, Zhou, Lin, et~al.]{yang2024qwen2}
An~Yang, Beichen Zhang, Binyuan Hui, Bofei Gao, Bowen Yu, Chengpeng Li, Dayiheng Liu, Jianhong Tu, Jingren Zhou, Junyang Lin, et~al.
\newblock Qwen2. 5-math technical report: Toward mathematical expert model via self-improvement.
\newblock \emph{arXiv preprint arXiv:2409.12122}, 2024.

\bibitem[Yang et~al.(2025)Yang, Li, Yang, Zhang, Hui, Zheng, Yu, Gao, Huang, Lv, et~al.]{yang2025qwen3}
An~Yang, Anfeng Li, Baosong Yang, Beichen Zhang, Binyuan Hui, Bo~Zheng, Bowen Yu, Chang Gao, Chengen Huang, Chenxu Lv, et~al.
\newblock Qwen3 technical report.
\newblock \emph{arXiv preprint arXiv:2505.09388}, 2025.

\bibitem[Yang et~al.(2022)Yang, Schuurmans, Abbeel, and Nachum]{yang2022chain}
Mengjiao~Sherry Yang, Dale Schuurmans, Pieter Abbeel, and Ofir Nachum.
\newblock Chain of thought imitation with procedure cloning.
\newblock \emph{Advances in Neural Information Processing Systems}, 35:\penalty0 36366--36381, 2022.

\bibitem[Yao et~al.(2023{\natexlab{a}})Yao, Yu, Zhao, Shafran, Griffiths, Cao, and Narasimhan]{yao2023tree}
Shunyu Yao, Dian Yu, Jeffrey Zhao, Izhak Shafran, Tom Griffiths, Yuan Cao, and Karthik Narasimhan.
\newblock Tree of thoughts: Deliberate problem solving with large language models.
\newblock \emph{Advances in Neural Information Processing Systems}, 36, 2023{\natexlab{a}}.

\bibitem[Yao et~al.(2023{\natexlab{b}})Yao, Xu, and Liu]{yao2023large}
Yuanshun Yao, Xiaojun Xu, and Yang Liu.
\newblock Large language model unlearning.
\newblock \emph{arXiv preprint arXiv:2310.10683}, 2023{\natexlab{b}}.

\bibitem[Ye et~al.(2024)Ye, Gao, Gong, Zheng, Jiang, Li, and Kong]{ye2024beyond}
Jiacheng Ye, Jiahui Gao, Shansan Gong, Lin Zheng, Xin Jiang, Zhenguo Li, and Lingpeng Kong.
\newblock Beyond autoregression: Discrete diffusion for complex reasoning and planning.
\newblock \emph{arXiv preprint arXiv:2410.14157}, 2024.

\bibitem[Yuan et~al.(2023{\natexlab{a}})Yuan, Yuan, Tan, Wang, Huang, and Huang]{yuan2023rrhf}
Hongyi Yuan, Zheng Yuan, Chuanqi Tan, Wei Wang, Songfang Huang, and Fei Huang.
\newblock Rrhf: Rank responses to align language models with human feedback.
\newblock \emph{Advances in Neural Information Processing Systems}, 36:\penalty0 10935--10950, 2023{\natexlab{a}}.

\bibitem[Yuan et~al.(2023{\natexlab{b}})Yuan, Yuan, Li, Dong, Lu, Tan, Zhou, and Zhou]{yuan2023scaling}
Zheng Yuan, Hongyi Yuan, Chengpeng Li, Guanting Dong, Keming Lu, Chuanqi Tan, Chang Zhou, and Jingren Zhou.
\newblock Scaling relationship on learning mathematical reasoning with large language models.
\newblock \emph{arXiv preprint arXiv:2308.01825}, 2023{\natexlab{b}}.

\bibitem[Yue et~al.(2025)Yue, Chen, Lu, Zhao, Wang, Song, and Huang]{yue2025does}
Yang Yue, Zhiqi Chen, Rui Lu, Andrew Zhao, Zhaokai Wang, Shiji Song, and Gao Huang.
\newblock Does reinforcement learning really incentivize reasoning capacity in llms beyond the base model?
\newblock \emph{arXiv preprint arXiv:2504.13837}, 2025.

\bibitem[Zelikman et~al.(2022)Zelikman, Wu, Mu, and Goodman]{zelikman2022star}
Eric Zelikman, Yuhuai Wu, Jesse Mu, and Noah Goodman.
\newblock Star: Bootstrapping reasoning with reasoning.
\newblock \emph{Advances in Neural Information Processing Systems}, 35:\penalty0 15476--15488, 2022.

\bibitem[Zelikman et~al.(2024)Zelikman, Harik, Shao, Jayasiri, Haber, and Goodman]{zelikman2024quiet}
Eric Zelikman, Georges Harik, Yijia Shao, Varuna Jayasiri, Nick Haber, and Noah~D Goodman.
\newblock Quiet-star: Language models can teach themselves to think before speaking.
\newblock \emph{arXiv preprint arXiv:2403.09629}, 2024.

\bibitem[Zhang et~al.(2024{\natexlab{a}})Zhang, Zhoubian, Hu, Yue, Dong, and Tang]{zhang2024rest}
Dan Zhang, Sining Zhoubian, Ziniu Hu, Yisong Yue, Yuxiao Dong, and Jie Tang.
\newblock Rest-mcts*: Llm self-training via process reward guided tree search.
\newblock \emph{arXiv preprint arXiv:2406.03816}, 2024{\natexlab{a}}.

\bibitem[Zhang et~al.(2024{\natexlab{b}})Zhang, Lin, Bai, and Mei]{zhang2024negative}
Ruiqi Zhang, Licong Lin, Yu~Bai, and Song Mei.
\newblock Negative preference optimization: From catastrophic collapse to effective unlearning.
\newblock \emph{arXiv preprint arXiv:2404.05868}, 2024{\natexlab{b}}.

\bibitem[Zhang et~al.(2024{\natexlab{c}})Zhang, Du, Pang, Liu, Gao, and Lin]{zhang2024chain}
Xuan Zhang, Chao Du, Tianyu Pang, Qian Liu, Wei Gao, and Min Lin.
\newblock Chain of preference optimization: Improving chain-of-thought reasoning in llms.
\newblock \emph{Advances in Neural Information Processing Systems}, 37:\penalty0 333--356, 2024{\natexlab{c}}.

\bibitem[Zhang et~al.(2024{\natexlab{d}})Zhang, Yang, Ke, Cui, Zheng, Wang, and Huang]{zhang2024safe}
Zhexin Zhang, Junxiao Yang, Pei Ke, Shiyao Cui, Chujie Zheng, Hongning Wang, and Minlie Huang.
\newblock Safe unlearning: A surprisingly effective and generalizable solution to defend against jailbreak attacks.
\newblock \emph{arXiv preprint arXiv:2407.02855}, 2024{\natexlab{d}}.

\bibitem[Zhao et~al.(2023)Zhao, Joshi, Liu, Khalman, Saleh, and Liu]{zhao2023slic}
Yao Zhao, Rishabh Joshi, Tianqi Liu, Misha Khalman, Mohammad Saleh, and Peter~J Liu.
\newblock Slic-hf: Sequence likelihood calibration with human feedback.
\newblock \emph{arXiv preprint arXiv:2305.10425}, 2023.

\bibitem[Zhao et~al.(2024)Zhao, Lee, and Hsu]{zhao2024large}
Zirui Zhao, Wee~Sun Lee, and David Hsu.
\newblock Large language models as commonsense knowledge for large-scale task planning.
\newblock \emph{Advances in Neural Information Processing Systems}, 36, 2024.

\bibitem[Zheng et~al.(2024)Zheng, Mishra, Zhang, Chen, Chen, Nova, Hou, Cheng, Le, Chi, et~al.]{zheng2024natural}
Huaixiu~Steven Zheng, Swaroop Mishra, Hugh Zhang, Xinyun Chen, Minmin Chen, Azade Nova, Le~Hou, Heng-Tze Cheng, Quoc~V Le, Ed~H Chi, et~al.
\newblock Natural plan: Benchmarking llms on natural language planning.
\newblock \emph{arXiv preprint arXiv:2406.04520}, 2024.

\bibitem[Zhou et~al.(2023)Zhou, Yan, Shlapentokh-Rothman, Wang, and Wang]{zhou2023language}
Andy Zhou, Kai Yan, Michal Shlapentokh-Rothman, Haohan Wang, and Yu-Xiong Wang.
\newblock Language agent tree search unifies reasoning acting and planning in language models.
\newblock \emph{arXiv preprint arXiv:2310.04406}, 2023.

\bibitem[Zhuang et~al.(2023)Zhuang, Chen, Yu, Mitra, Bursztyn, Rossi, Sarkhel, and Zhang]{zhuang2023toolchain}
Yuchen Zhuang, Xiang Chen, Tong Yu, Saayan Mitra, Victor Bursztyn, Ryan~A Rossi, Somdeb Sarkhel, and Chao Zhang.
\newblock Toolchain*: Efficient action space navigation in large language models with a* search.
\newblock \emph{arXiv preprint arXiv:2310.13227}, 2023.

\bibitem[Zhuo et~al.(2024)Zhuo, Vu, Chim, Hu, Yu, Widyasari, Yusuf, Zhan, He, Paul, et~al.]{zhuo2024bigcodebench}
Terry~Yue Zhuo, Minh~Chien Vu, Jenny Chim, Han Hu, Wenhao Yu, Ratnadira Widyasari, Imam Nur~Bani Yusuf, Haolan Zhan, Junda He, Indraneil Paul, et~al.
\newblock Bigcodebench: Benchmarking code generation with diverse function calls and complex instructions.
\newblock \emph{arXiv preprint arXiv:2406.15877}, 2024.

\end{thebibliography}
